\documentclass[10pt,twocolumn,letterpaper]{article}

\usepackage[pagenumbers]{iccv} 

\usepackage{epsfig}
\usepackage{bbding}

\usepackage{graphicx}
\usepackage{booktabs}
\usepackage{amsmath}
\usepackage{amssymb}

\usepackage{tabularx}
\usepackage{multicol}
\usepackage{multirow}

\usepackage{subcaption}
\usepackage{soul}
\usepackage[normalem]{ulem}
\usepackage{algorithm}
\usepackage{algorithmic}
\usepackage{mathtools}
\usepackage{bm}
\usepackage{color, colortbl}
\usepackage{stfloats}
\usepackage{enumitem}  
\usepackage{siunitx}[=v2]

\definecolor{LightGreen}{rgb}{0.9,1,0.9}
\definecolor{LightGray}{rgb}{0.92,0.92,0.92}
\definecolor{LightBlue}{rgb}{0.8,0.9,1.0}

\definecolor{darkgreen}{RGB}{0,128,0}
\definecolor{darkred}{RGB}{139,0,0}

\newcommand{\increase}[1]{\textcolor{darkred}{{$\uparrow$#1\%}}}
\newcommand{\reduce}[1]{\textcolor{darkgreen}{{$\downarrow$#1\%}}}

\definecolor{iccvblue}{rgb}{0.21,0.49,0.74}
\usepackage[pagebackref,breaklinks,colorlinks,allcolors=iccvblue]{hyperref}

\def\paperID{****} 
\def\confName{ICCV}
\def\confYear{2025}

\newcommand{\methodname}{UniGaze\xspace}

\title{\methodname: Towards Universal Gaze Estimation via Large-scale Pre-Training}

\author{Jiawei Qin$^{1}$, Xucong Zhang$^{2}$, Yusuke Sugano$^{1}$ \\
\normalsize $^1$ Institute of Industrial Science, The University of Tokyo, Komaba 4-6-1, Tokyo, Japan \\
\normalsize $^2$ Computer Vision Lab, Delft University of Technology, Mekelweg 5, Delft, Netherlands \\
{
\tt\small \{jqin, sugano\}@iis.u-tokyo.ac.jp}\\
\tt\small xucong.zhang@tudelft.nl
}

\begin{document}

\twocolumn[{%
\renewcommand\twocolumn[1][]{#1}%
\maketitle
\vspace{-25pt}
\begin{center}
  \includegraphics[width=0.9\linewidth]{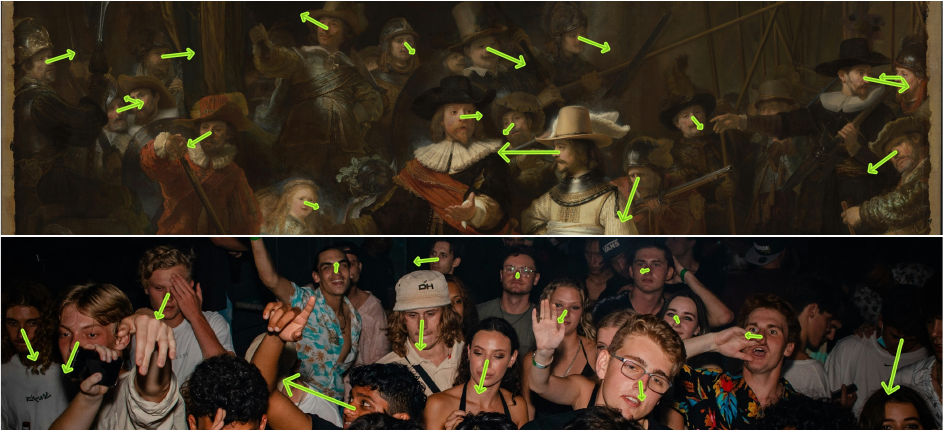}
  \captionof{figure}{
  Leveraging self-supervised pre-training on large-scale facial data, the proposed \methodname demonstrates strong generalization to unseen in-the-wild face images under diverse conditions, including facial appearance, lighting conditions, variant head poses, and face resolutions. We draw the estimated gaze direction with green arrows. More examples are shown in the supplementary materials.
  }
  \label{fig:teaser}
\end{center}%
}]

\begin{abstract}
Despite decades of research on data collection and model architectures, current gaze estimation models encounter significant challenges in generalizing across diverse data domains. 
Recent advances in self-supervised pre-training have shown remarkable performances in generalization across various vision tasks.
However, their effectiveness in gaze estimation remains unexplored.
We propose \methodname, for the first time, leveraging large-scale in-the-wild facial datasets for gaze estimation through self-supervised pre-training.
Through systematic investigation, we clarify critical factors that are essential for effective pre-training in gaze estimation. 
Our experiments reveal that self-supervised approaches designed for semantic tasks fail when applied to gaze estimation, while our carefully designed pre-training pipeline consistently improves cross-domain performance. 
Through comprehensive experiments of challenging cross-dataset evaluation and novel protocols including leave-one-dataset-out and joint-dataset settings, we demonstrate that \methodname significantly improves generalization across multiple data domains while minimizing reliance on costly labeled data. 
Source code and model are available at \url{https://github.com/ut-vision/UniGaze}.
\end{abstract}

\section{Introduction}
\label{sec:intro}

Gaze estimation is a key task in computer vision with broad applications in human-computer interaction~\cite{majaranta2014eye,belardinelli2024gaze}, virtual reality~\cite{lystbaek2022gaze}, and behavioral analysis~\cite{harezlak2018application,ke2024using,sharma2024review}. 
The methodology in unconstrained environments has been actively researched in past decades~\cite{baluja1993non,tan2002appearance,lu2012head,lu2014learning,zhang2015appearance,cheng2024appearance,ghosh2023automatic,wang2019generalizing}.
However, achieving robust and accurate gaze estimation in unseen test environments remains a fundamental challenge.
Although current models can achieve high accuracy when tailored to specific datasets or individuals~\cite{park2019few}, they consistently show significant performance degradation in novel environments. 
This challenge is evidenced by the persistent difficulty in training one single model to achieve high accuracy across various gaze estimation datasets~\cite{funes2014eyediap,mpii_zhang19_pami,krafka2016eye,kellnhofer2019gaze360,zhang2020eth}.
Despite decades of research on data collection and model architecture, the problem of generalization across variations in head pose, identity, and lighting conditions remains largely unsolved.

Recent advances in self-supervised pre-training have shown remarkable potential for improving model generalization to arbitrary data domains.
Following scaling laws~\cite{kaplan2020scaling}, larger models trained with extensive data and compute resources can be more sample-efficient, which has been observed with significant performance improvements across numerous tasks such as image classification~\cite{srivastava2024omnivec,singh2023effectiveness,fang2023eva}, segmentation~\cite{li2023dreamteacher,xiang2023denoising,zhao2023unleashing,kirillov2023segment}, human-centric tasks~\cite{khirodkar2025sapiens}, gaze following~\cite{lin2024gazehta}, and face-focused applications~\cite{zheng2022general,cai2023marlin,gao2024self,wang2023toward,sun2024face}.  
However, their effectiveness in gaze estimation remains unexplored.
Unlike these typical tasks, gaze estimation is a geometric regression task to map facial appearance to precise 3D vectors.
Our experiments show that existing pre-trained models~\cite{zheng2022general,caron2021emerging,chen2021empirical} fail to improve gaze estimation, as they are optimized for semantic understanding rather than the fine-grained facial structure crucial for gaze direction.
This raises an essential question: Can large-scale pre-training benefit the fine-grained geometric nature of gaze estimation? 
Our research offers an answer: Yes, but \textit{only when the pre-training is specifically tailored to the unique constraints of gaze estimation}.

In this work, we present a novel approach toward \textbf{Uni}versal \textbf{Gaze} estimation, dubbed \methodname, exploring the potential of large-scale self-supervised pre-training for appearance-based gaze estimation. 
We employ Masked Autoencoder (MAE)~\cite{he2022masked} on the Vision Transformer (ViT) architecture~\cite{dosovitskiy2020vit}, using diverse in-the-wild face image datasets.
Through systematic experimentation, we discovered that pre-training for gaze estimation requires three essential components that differ from typical pre-training approaches: 
(1) pre-training on normalized facial images maintaining the spatial configuration required by downstream gaze models~\cite{zhang2018revisiting}, 
(2) ensuring diverse yet balanced head pose distributions to learn robust facial representations across viewing angles, and 
(3) incorporating sufficient identity diversity to generalize across different facial appearances.
This strategy allows the model to learn appropriate feature representations within the specific input space required by gaze estimation models, enabling effective transfer to the downstream gaze estimation task.

Through extensive experiments, we demonstrate that training our pre-trained \methodname on gaze-specific datasets yields substantial generalization performance improvements across multiple data domains~\cite{zhang2020eth,mpii_zhang19_pami,funes2014eyediap,krafka2016eye,kellnhofer2019gaze360}, surpassing state-of-the-art domain generalization methods~\cite{cheng2022puregaze,xu2023learning,bao2024feature,yin2024clip,yin2024lggaze,zhao2024improving}.
Pre-training with general semantics~\cite{zhou2024deformable,caron2021emerging,chen2021empirical}, face-specific semantics~\cite{zheng2022general}, and vanilla MAE trained on small-scale gaze data~\cite{jiang2024learning} all fail to transfer effectively to gaze estimation, sometimes even performing worse than simple CNNs.
In contrast, our carefully designed pre-training approach consistently improves accuracy across diverse environments.
This highlights our crucial discovery that suitable datasets, normalized facial images, and varied yet balanced head pose distributions are vital for developing transferable representations for gaze estimation, offering practical guidelines for future research in this field.

Additionally, in gaze estimation, the widely used cross-dataset evaluation usually only trains the model on a single dataset, which cannot fully represent the appearance diversity and pose range in real-world environments.
There also remains the possibility that some adaptation methods merely overfit some specific training/testing dataset combinations.
To address this limitation and reflect the practical requirements of real-world applications, we propose two novel evaluation protocols utilizing multiple datasets for training: a \textit{leave-one-dataset-out} setting that assesses generalization to unseen datasets and a \textit{joint-dataset} setting that evaluates the achievable performance across multiple datasets.
Our comprehensive evaluation demonstrates that \methodname consistently achieves superior performance across diverse environments under these protocols, suggesting the effectiveness of large-scale pre-training.

In summary, our contributions are threefold: 
(i) We present \methodname, a novel gaze estimation model that addresses the fundamental challenge of cross-domain generalization in appearance-based gaze estimation via large-scale pre-training.
(ii) We provide empirical evidence that MAE pre-training on normalized face images can learn meaningful representations for gaze estimation tasks.
(iii) We propose leave-one-dataset-out and joint-dataset evaluation protocols that offer practical benchmarks for assessing gaze estimation performance in real-world scenarios.

\section{Related Works}
\label{sec:related}

\subsection{Appearance-based Gaze Estimation}
Appearance-based methods for gaze estimation have gained prominence, leveraging the ability of deep learning to learn gaze representations from gaze-labeled datasets~\cite{zhang2015appearance,mpii_zhang19_pami,zhang2020eth,kellnhofer2019gaze360,krafka2016eye,funes2014eyediap,fischer2018rt}.
However, these approaches are limited by the scarcity of comprehensive, well-labeled gaze datasets, which hinders performance in unconstrained settings.
One solution to data scarcity has been synthetic data generation, where gaze images are synthesized with controllable variables such as lighting, head pose, and redirected gaze~\cite{qin2022learning,zheng2020self,jin2023redirtrans,ruzzi2023gazenerf,wang2023high_nerf,yin2022nerfgaze}. 
Despite its utility, synthetic data often lacks realism, leading to domain adaptation challenges~\cite{wood2015rendering,kim2019nvgaze}. 
Furthermore, subtle inaccuracies in gaze labels can compromise model performance when transferred to real-world scenarios.

In addition to data-driven approaches, method-driven solutions have been explored to improve robustness and generalizability~\cite{yin2024clip,liu2024gaze,bao2024feature,cheng2022puregaze,zhao2024improving,xu2023learning}.
For instance, gaze frontalization~\cite{xu2024gaze}, multi-view consistency~\cite{hisadome2024rotation,bao2024unsupervised}, and contrastive learning~\cite{he2020momentum,jindal2023contrastive} have been used to learn generalized gaze representations.
Clip-Gaze~\cite{yin2024clip} and LG-Gaze~\cite{yin2024lggaze} use the linguistic features extracted from the vision-language model to regularize the gaze feature learning.
Alternatively, Kothari~\etal~\cite{kothari2021weakly} utilizes in-the-wild face datasets with \textit{look-at-each-other} labels as weak supervision.
Furthermore, domain adaptation methods~\cite{park2019few,wang2022contrastive} incorporate target domain samples with limited or no labels.
For example, PnP-GA+ uses the plug-and-play method to adapt the gaze estimation model to new domains with assembling model variants~\cite{liu2024pnp}.

Network architectures in gaze estimation remain predominantly CNN-based, particularly relying on ResNet~\cite{he2016deep}.
Recent work by Cheng~\etal~\cite{cheng2022gaze} explored ViTs~\cite{dosovitskiy2020vit} for gaze estimation, finding that although ViTs hold promise, they require extensive pre-training data beyond standard ImageNet~\cite{deng2009imagenet} to perform effectively.
This motivates the exploration of ViT models specifically tailored to learn diverse gaze representations through pre-training approaches.

\subsection{Large-scale Pre-Training in Vision Models}\label{sec:related_pretrain}
Large-scale pre-training has become fundamental for foundational model development in computer vision. 
Recent studies show that pre-trained generative models enhance representation learning across various applications. 
For example, diffusion models~\cite{ho2020denoising,rombach2022high} have been successfully applied to image classification~\cite{li2023your} and segmentation~\cite{li2023dreamteacher,xiang2023denoising,zhao2023unleashing}.
Another effective unsupervised pre-training approach is masked autoencoding~\cite{he2022masked,srivastava2024omnivec,singh2023effectiveness,fang2023eva}, which demonstrates strong capabilities in image recognition and robust feature extraction.
It is also shown that masked image modeling benefits more for ViT than CNN family~\cite{fang2022corrupted,kraus2024masked}.

MAE pre-training has been adapted to several domains such as audio~\cite{huang2022masked}, video~\cite{tong2022videomae}, and microscopy~\cite{kraus2024masked}.
Recently, Sapiens~\cite{khirodkar2025sapiens} leverages human-centric data for MAE pre-training, resulting in strong generalization across multiple human-related tasks.
These adaptations to specific domains have made a crucial step in advancing research.
Similarly, multiple self-supervised pre-training works have been proposed to achieve notable improvement in face-centric tasks, such as expression recognition, facial attribute recognition, and face alignment~\cite{zheng2022general,cai2023marlin,gao2024self,wang2023toward,sun2024face}.
These studies underscore the effectiveness of pre-training masked autoencoders on large, diverse datasets for enhancing model robustness and generalizability.
In contrast, gaze estimation has received less attention in this context.
A concurrent work~\cite{jiang2024learning} even highlighted a limitation in applying MAE to gaze estimation with ViT, noting that random masking tends to focus on global semantics while neglecting critical gaze-related information.

\section{\methodname}\label{sec:method}

Our \methodname model consists of a large-scale pre-training stage followed by task-specific training. 
The pre-training stage utilizes a carefully curated dataset of \SI{1.6}{\mega{}} facial images that combines real-world and synthetic data to ensure broad coverage of head poses and facial appearances. 
We adopt the MAE framework to learn robust facial representations from this diverse dataset and then train it with labeled gaze data for the downstream gaze estimation task. 

\subsection{Pre-Training Datasets}\label{sec:pretrain_data_prepare}

\begin{table}[t]
    \begin{center}
        \resizebox{0.94\linewidth}{!}{ 
        \setlength{\tabcolsep}{6pt}
            \begin{tabular}{l|cc|c}
            \toprule
            \textbf{Dataset}  & \textbf{Type} & \textbf{\# Identities} & \textbf{\# Samples} \\
            \midrule
            CelebV-Text~\cite{yu2023celebv} & Real & 13,179 & 666,967 \\
            VFHQ~\cite{xie2022vfhq} & Real & 10,382 & 231,809 \\
            VGGFace2~\cite{cao2018vggface2} & Real & 9,131 & 182,603 \\
            \midrule
            FaceSynthetics~\cite{wood2021fake} & Syn. & 86,878$^{\dagger}$ & 86,878 \\
            SFHQ-T2I~\cite{david_beniaguev_2024_SFHQ_T2I} & Syn. & 120,241$^{\dagger}$ & 120,241 \\
            FFHQ-NV~\cite{Karras2019stylegan2, qin2022learning} & Syn. & 25,000 & 100,000 \\
            XGaze-Dense~\cite{zhang2020eth,agisoft_metashape,qin2023domain} & Syn. & 60 & 267,160 \\
            \midrule
            Total  & - & 264,871 & 1,655,668\\
            \bottomrule
            \end{tabular}
        }
    \end{center}
    \caption{
        Statistics of face datasets used to pre-train \methodname in terms of data type to be real or synthetic (Syn.), number of identities, and number of samples.
        The $^{\dagger}$ indicates that we assume there are no duplicated identities during the synthesis image generation.
    }\label{table:face_data_info}
\end{table}

Learning robust gaze representations requires extensive head pose and facial appearance variations. 
To achieve this, we combine two complementary data sources: real data that captures natural face appearances and synthetic data that allows us to cover extreme pose variations and diverse appearances systematically.

\Cref{table:face_data_info} summarizes our pre-training dataset composition, which comprises approximately \SI{1.6}{\mega{}} samples.
This combined dataset includes over \SI{260}{\kilo{}} unique identities, vastly exceeding the 1,474 subjects in the existing GazeCapture dataset~\cite{krafka2016eye}. 
Although GazeCapture~\cite{krafka2016eye} and ETH-XGaze~\cite{zhang2020eth} offer over \SI{2.4}{\mega{}} and \SI{1}{\mega{}} samples, respectively, our combined dataset provides a significantly higher level of diversity.
In this manner, our pre-training data offers a significantly broader representation regarding facial appearances, head poses, and environmental conditions, surpassing existing gaze estimation datasets.

\begin{figure*}[t]
  \begin{center}
      \includegraphics[width=0.9\linewidth]{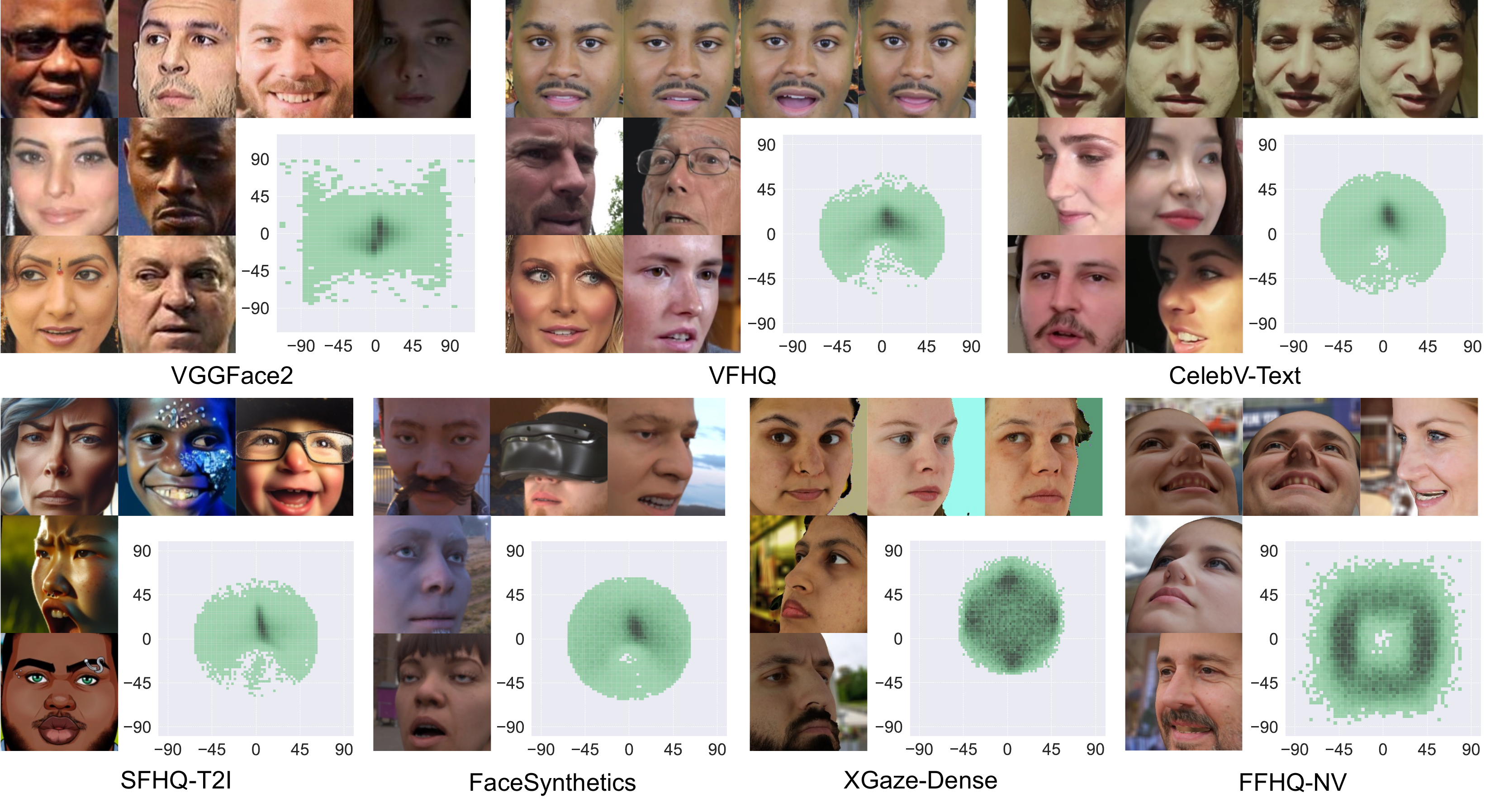}
      \caption{Example of the normalized facial images from different datasets in the pre-training stage. We also draw their head pose distributions where the vertical axis is the pitch rotation angle and the horizontal axis is the yaw rotation angle in degrees.
      }\label{fig:face_samples_distribution}
  \end{center}
\end{figure*}

\noindent\textbf{Real Datasets.} 
We use VGGFace2~\cite{cao2018vggface2} for its large number of identities with diverse conditions. 
Additionally, we incorporate two high-quality video datasets, VFHQ~\cite{xie2022vfhq} and CelebV-Text~\cite{yu2023celebv}, leveraging their natural pose and gaze variations provided by video sequences. 
To reduce redundancy while maintaining diversity, we sub-sample every 15 frames from VFHQ, 45 frames from CelebV-Text, and select 20 images per identity from VGGFace2.

\noindent\textbf{Synthetic Datasets.} 
To ensure diverse facial appearances, we use two synthetic datasets: FaceSynthetics~\cite{wood2021fake}, generated via computer graphics, and SFHQ-T2I~\cite{david_beniaguev_2024_SFHQ_T2I}, created with diffusion models. 
We further employ novel-view synthesis techniques to extend the range in head poses.
We reconstruct 3D facial shapes from FFHQ~\cite{Karras2019stylegan2} via the single-view method~\cite{qin2022learning}, synthesizing FFHQ-NV. 
Following~\cite{qin2023domain}, we use Metashape~\cite{agisoft_metashape} for multi-view 3D face reconstruction and apply novel-view rendering to get XGaze-Dense. 
Since XGaze-Dense is used without gaze labels, its role is equivalent to that of a generic facial dataset in our pre-training.

\noindent\textbf{Data Pre-processing}
We perform facial landmark detection~\cite{bulat2017far} to estimate the 3D head pose by perspective-n-point (PnP) algorithm~\cite{fischler1981random}.
We then apply data normalization~\cite{zhang2018revisiting} to crop face images, ensuring alignment of the input space between MAE pre-training and gaze estimation. 
We filter out samples with extreme head poses for VFHQ, CelebV-Text, SFHQ-T2I, and FaceSynthetics to eliminate extreme cases where the face is invisible. 
Precisely, we discard samples with an $L_2$ norm of pitch and yaw angles exceeding 80 degrees.
\Cref{fig:face_samples_distribution} shows example face images and head pose distributions for each dataset.

\subsection{Training Procedure}

We follow the MAE~\cite{he2022masked} to pre-train the ViT model without any gaze labels. 
Briefly, it randomly masks patches of the input image and the model is trained to predict the masked contents.
Given an input image $\bm{X}$, it is divided into $N$ patches $\{x_i\}_{i=1}^N$.
A subset of these patches is masked out, denoted by $\{x_i\}_{i \in M}$ with $M \subset \{1, \ldots, N\}$.
The encoder processes the visible patches $\{x_i\}_{i \in V}$, where $V = \{1, \ldots, N\} \setminus M$, and generates a latent representation $z$.
An extra decoder then takes $z$ and reconstructs the masked patches $\{\hat{x}_i\}_{i \in M}$ with the loss $\mathcal{L}_{\text{MAE}} = \frac{1}{|M|} \sum_{i \in M} \| x_i - \hat{x}_i \|^2$.
This process encourages the model to capture meaningful structures and features in $z$, making it well-suited for downstream tasks~\cite{khirodkar2025sapiens}.
During pre-training, we randomly apply flip, central crop, and color jitter, and the mask ratio is 75\%.
The loss is computed with the pixel value normalized within each patch, which is suggested to have better representation~\cite{he2022masked}.

\noindent\textbf{Gaze Estimation Training}
With labeled gaze datasets, we train the pre-trained model for the gaze estimation downstream task. 
We replace the decoder with a fully connected layer to predict gaze direction from the latent representation $z$.
The gaze direction is represented by a 2D vector in the polar angle coordinate system.
We use the loss $\mathcal{L}_{1} = \| \bm{g} - \hat{\bm{g}} \|_1$, where $\bm{g}$ and $\hat{\bm{g}}$ denote the ground-truth gaze label and the prediction, respectively.

\subsection{Implementation Details}
We use the Adam optimizer~\cite{kingma2014adam} with a base learning rate of \num{1.5e-4} and a weight decay of \num{0.05}.
We set a batch size of 4,096 and pre-train the ViT-Huge model for 300 epochs, which requires approximately 120 hours on four NVIDIA H100 GPUs.
During gaze estimation training, we do not apply any image augmentation.
We use the Adam optimizer~\cite{kingma2014adam} with a learning rate of \num{1e-4} and a weight decay of \num{1e-6}, and the one-cycle learning rate~\cite{smith2019super}.
More details and examples can be found in supplementary materials.

\begin{table*}[t]
\begin{center}
    \begin{subtable}[t]{0.33\textwidth}
    \begin{center}
    \resizebox{0.99\linewidth}{!}{ 
    \setlength{\tabcolsep}{4.5pt}
        \begin{tabular}{l|cccc}
        \toprule
        \textbf{Models \textbackslash~Test}  & \textbf{M} & \textbf{GC} & \textbf{E} & \textbf{G360} \\
        \midrule
        ResNet-50        & 6.75 & 10.08 & 10.28 & 19.80 \\
        GazeTR-50 & 7.09 & 10.95 & 9.47 & 21.10 \\   
        \midrule
        DINO-B~\cite{caron2021emerging} & 7.73 & 9.12 & 11.98 & 19.44 \\
        MoCo-v3-B~\cite{chen2021empirical} & 7.19 & 8.88 & 9.50 & 19.09 \\
        FaRL-B~\cite{zheng2022general} & 7.18& 8.56& 9.53&17.28 \\
        \rowcolor{LightGray}
        \methodname-B & 6.21 & 7.35 & 6.64 & 12.18 \\
       \midrule
        ViT-H   & 7.68 & 9.36 & 10.40& 20.38 \\
       \rowcolor{LightGray}
      \methodname-H  & \textbf{5.57} & \textbf{6.56}& \textbf{6.53} & \textbf{11.19} \\
        \bottomrule
        \end{tabular}
    }
    \caption{ Results when trained on XGaze. }
    \label{table:cross_x}
    \end{center}
    \end{subtable}
    \vspace{8pt} 
    \begin{subtable}[t]{0.33\textwidth}
    \begin{center}
    \resizebox{0.99\linewidth}{!}{ 
    \setlength{\tabcolsep}{4.5pt}
       \begin{tabular}{l|cccc}
        \toprule
        \textbf{Models \textbackslash~Test} & \textbf{X\textsubscript{Test}} & \textbf{GC} & \textbf{E} & \textbf{G360} \\
        \midrule
        ResNet-50        & 32.80  & 6.75 & 17.11 & 27.92 \\
        GazeTR-50 & 29.00 & 7.06 & 19.38 & 28.22 \\
        \midrule
        DINO-B~\cite{caron2021emerging} & 33.41 & 8.13 & 20.07 & 30.39 \\
        MoCo-v3-B~\cite{chen2021empirical} & 30.99&6.53&17.56&27.14\\
        FaRL-B~\cite{zheng2022general} &30.43 &6.14 &16.55 &25.97 \\
        \rowcolor{LightGray}
         \methodname-B & 30.56 & 5.68 & 17.54 & \textbf{20.85} \\
        \midrule
        ViT-H   & \textbf{28.25} & 6.87 & 16.02 & 25.30 \\
        \rowcolor{LightGray}
        \methodname-H      & 33.11 &\textbf{4.87}&\textbf{12.66}&21.28 \\
        \bottomrule
        \end{tabular}
    }
    \caption{ Results when trained on MPIIFaceGaze.}
    \label{table:cross_m}
    \end{center}
    \end{subtable}
    \begin{subtable}[t]{0.33\textwidth}
    \begin{center}
    \resizebox{0.99\linewidth}{!}{ 
    \setlength{\tabcolsep}{4.5pt}
       \begin{tabular}{l|cccc}
        \toprule
        \textbf{Models \textbackslash~Test} & \textbf{X\textsubscript{Test}} & \textbf{M} & \textbf{E} & \textbf{G360} \\
        \midrule
        ResNet-50        & 26.56 & 5.84 & 13.39 & 25.33 \\
        GazeTR-50 & 23.57&5.49 &14.25 &25.48 \\
        \midrule
        DINO-B~\cite{caron2021emerging} & 27.56 & 6.97 & 16.75 & 26.49 \\
        MoCo-v3-B~\cite{chen2021empirical} &27.26&5.29&14.75&25.94\\
        FaRL-B~\cite{zheng2022general} & 26.55 &5.74&15.18&23.49 \\
        \rowcolor{LightGray}
         \methodname-B & \textbf{20.63} & 5.01 & 10.91 & 18.91 \\
        \midrule
        ViT-H   & 23.49 & 5.54 & 14.60 & 23.98\\
        \rowcolor{LightGray}
        \methodname-H  & 22.67&\textbf{4.89}&\textbf{10.61}&\textbf{17.77} \\
        \bottomrule
        \end{tabular}
    }
    \caption{ Results when trained on GazeCapture.}
    \label{table:cross_gc}
    \end{center}
    \end{subtable}
    \\
    \begin{subtable}[t]{0.49\textwidth}
    \begin{center}
    \resizebox{0.75\linewidth}{!}{ 
    \setlength{\tabcolsep}{6pt}
       \begin{tabular}{l|cccc}
        \toprule
        \textbf{Models \textbackslash~Test} & \textbf{X\textsubscript{Test}} & \textbf{M} & \textbf{GC} & \textbf{G360} \\
        \midrule
        ResNet-50        & 37.50 & 16.88 & 16.37 & 30.03 \\
        GazeTR-50 & 35.82 & 16.37 &16.63& 25.69 \\
        \midrule
        DINO-B~\cite{caron2021emerging} & 38.21 & 17.84 & 18.45 & 33.49 \\
        MoCo-v3-B~\cite{chen2021empirical} &31.98&14.29&14.15&24.90\\
        FaRL-B~\cite{zheng2022general} & 34.74 &12.96&12.31&26.88 \\
        \rowcolor{LightGray}
         \methodname-B &25.37 & \textbf{8.91} & \textbf{8.89} & 18.27 \\
        \midrule
        ViT-H   & 30.90 & 15.51& 13.51 &26.12 \\
        \rowcolor{LightGray}
        \methodname-H   &\textbf{23.79}&8.93&9.97&\textbf{16.00} \\
        \bottomrule
        \end{tabular}
    }
    \caption{ Results when trained on EYEDIAP.}
    \label{table:cross_ed}
    \end{center}
    \end{subtable}
    \begin{subtable}[t]{0.49\textwidth}
    \begin{center}
    \resizebox{0.75\linewidth}{!}{ 
    \setlength{\tabcolsep}{6pt}
       \begin{tabular}{l|cccc}
        \toprule
        \textbf{Models \textbackslash~Test} & \textbf{X\textsubscript{Test}} & \textbf{M} & \textbf{GC} & \textbf{E}  \\
        \midrule
        ResNet-50   & 18.83 & 10.25 & 9.90 & 12.06 \\
        GazeTR-50 & 18.04 & 11.28 & 10.82 & 13.69\\
        \midrule
        DINO-B~\cite{caron2021emerging} & 22.25 & 11.02 & 11.17 & 17.53 \\
        MoCo-v3-B~\cite{chen2021empirical} &17.59&7.50&8.72&11.91\\
        FaRL-B~\cite{zheng2022general} & 20.55 &7.62&7.75&12.43 \\
        \rowcolor{LightGray}
         \methodname-B & \textbf{12.22} & 6.00 & 8.11 & 8.74 \\
        \midrule
        ViT-H   & 18.63 & 8.24 & 9.43 & 11.50 \\
        \rowcolor{LightGray}
        \methodname-H   & 16.39 & \textbf{5.43} & \textbf{6.48}& \textbf{6.97}\\
        \bottomrule
        \end{tabular}
    }
    \caption{ Results when trained on Gaze360.}
    \label{table:cross_g360}
    \end{center}
    \end{subtable}
    \caption{
        Cross-dataset evaluation of different models trained on one dataset and tested on multiple unseen datasets. 
        Each subtable corresponds to a specific training dataset, with columns representing the testing datasets (\textbf{X\textsubscript{Test}}: XGaze Test, \textbf{M}: MPIIFaceGaze, \textbf{GC}: GazeCapture, \textbf{E}: EYEDIAP, \textbf{G360}: Gaze360). 
        Results demonstrate the generalization ability of each model, with \methodname consistently outperforming other baselines in most settings, showcasing the effectiveness of our pre-training approach.
        }
\label{table:cross_dataset}
\end{center}
\end{table*}


\section{Experiments}\label{sec:experiments}

\subsection{Experimental Settings}\label{sec:exp_setting}

\paragraph{Gaze Datasets}
We conduct experiments on multiple gaze estimation datasets following the common way of utilization in recent works.
\textbf{MPIIFaceGaze}~\cite{zhang2017s} contains 15 subjects with nearly frontal head poses. 
In experiments requiring splitting, the first ten subjects are used for training and the remaining five for testing.
\textbf{ETH-XGaze}~\cite{zhang2020eth} comprises over \SI{750}{\kilo{}} publicly available gaze-labeled images of 80 subjects. We refer to the ``XGaze'' as the 80 subjects, while ``XGaze Train/Test'' indicates 60/20-subject split.
When training, we randomly select three out of the 18 cameras to reduce redundancy without losing effectiveness, and we utilize all cameras for testing. 
\textbf{EYEDIAP}~\cite{funes2014eyediap} includes 16 subjects and two sessions with screen (CS) and 3D floating object (FT) targets. 
We use both sessions and split the data into training and test sets by subjects, with an 8/8 split.
The data is pre-processed using the pipeline by Park~\etal~\cite{park2019few}. 
\textbf{Gaze360}~\cite{kellnhofer2019gaze360} consists of indoor and outdoor images of 238 subjects with wide ranges of head poses and gaze directions. We use the training and test split defined in the original paper.
\textbf{GazeCapture}~\cite{krafka2016eye} contains around 1400 subjects collected through crowd-sourcing. We use the training and test split defined in the original GazeCapture paper and pre-process with the pipeline by Park~\etal~\cite{park2019few}.
When training, we sample every 15 frames to reduce redundancy, and we use all samples for testing.

\paragraph{Baseline Architectures}
We draw upon existing gaze estimation research and consider baselines of convolutional neural networks, ViTs, and hybrid models.
\textbf{ResNet}~\cite{he2016deep} models are lightweight yet powerful CNNs, which dominate the backbone and baseline in most of the gaze estimation works \cite{zhang2020eth,cheng2022puregaze,xu2023learning,zhao2024improving}. 
We use ResNet-50 in our experiments.
\textbf{GazeTR-50 (Hybrid)}~\cite{cheng2022gaze} is a hybrid network where the image features extracted from the ResNet-50 are fed into the transformer.
We include \textbf{DINO-B}~\cite{caron2021emerging}, \textbf{MoCo-v3-B}~\cite{chen2021empirical}, and \textbf{FaRL-B}~\cite{zheng2022general} as representative pre-trained ViT-Base models: DINO and MoCo-v3 have rich general semantic representations, while FaRL is specialized for face analysis tasks and pre-trained on \SI{20}{\mega{}} LAION-Face samples.
We also include the ImageNet~\cite{deng2009imagenet} pre-trained \textbf{ViTs}~\cite{dosovitskiy2020vit} for comparison.
We use the Base (B), Large (L), and Huge (H) variants in our experiments.

\begin{table}[t]
\begin{center}
    \resizebox{0.99\linewidth}{!}{ 
    \setlength{\tabcolsep}{4pt}
        \begin{tabular}{l|cccc}
        \toprule
        \textbf{Models}  & \textbf{X}$\rightarrow$\textbf{M} & \textbf{X}$\rightarrow$\textbf{E\textsubscript{CS}}& \textbf{G360}$\rightarrow$\textbf{M} & \textbf{G360}$\rightarrow$\textbf{E\textsubscript{CS}} \\
        \midrule
        \rowcolor{LightGray}
        ResNet-18$^{\dagger}$~\cite{zhao2024improving}  & 8.02 & 9.11 & 8.04 & 9.20 \\
        \rowcolor{LightGray}
        PureGaze$^{\dagger}$~\cite{cheng2022puregaze}  & 7.08 & 7.48 & 9.28 & 9.32 \\
        \rowcolor{LightGray}
        Gaze-Consistent$^{\dagger}$~\cite{xu2023learning} & 6.50 & 7.44 & 7.55 & 9.03 \\
        \rowcolor{LightGray}
        AGG$^{\dagger}$~\cite{bao2024feature}     & \underline{5.91} & 6.75 & 7.87  & 7.93 \\
        \rowcolor{LightGray}
        CLIP-Gaze$^{\dagger}$~\cite{yin2024clip}     & 6.41 & 7.51 & 6.89 & 7.06 \\
        \rowcolor{LightGray}
        LG-Gaze$^{\dagger}$~\cite{yin2024lggaze}     & 6.45 & 7.22 & 6.83 & 6.86 \\
        \rowcolor{LightGray}
        Gaze-BAR$^{\dagger}$~\cite{zhao2024improving}     & 6.35 & 6.72 & 6.96 & 8.79 \\
        ViT-H & 7.68 & 8.58 & 8.24 & 7.79 \\ 
        \midrule
        \methodname-B & 6.21 & \underline{5.08} & \underline{6.00} & \underline{6.63} \\
        \methodname-H & \textbf{5.57} &  \textbf{4.65} & \textbf{5.43} & \textbf{5.35} \\
        \bottomrule
        \end{tabular}

}
\end{center}
\caption{
    Domain generalization compared with SOTA methods in the cross-dataset setting. 
    Since most of the methods in the table are based on ResNet-18, we also show the ViT-H baseline that only pre-trained on ImageNet for fair comparison with \methodname-H.
}
\label{table:sota}
\end{table}

\begin{table}[t]
    \centering
    \resizebox{0.88\linewidth}{!}{ 
    \setlength{\tabcolsep}{4.5pt}
         \begin{tabular}{l|ccccc}
        \toprule
        \textbf{Methods \textbackslash~Dataset} & \textbf{X} & \textbf{M} & \textbf{GC} & \textbf{E} & \textbf{G360} \\
        \midrule
        Abdelrahman~\etal~\cite{abdelrahman2023l2cs}  & - & 3.92 & - & - & - \\
        Guan~\etal~\cite{guan2023end} & - & - & - & - & 9.81 \\  
        Shi~\etal~\cite{shi2024agent} & - & \textbf{3.61} & - & 4.78 & - \\
        3DGazeNet~\cite{ververas20253dgazenet} & 4.2 & 4.0 & 3.1 & - & 9.6 \\
        \midrule
        \methodname-H & \textbf{3.96} & 4.07 & \textbf{3.01} & \textbf{4.34} & \textbf{9.44} \\
        \bottomrule
        \end{tabular}
        }
    \caption{
    Within-dataset evaluation compared with SOTA methods. Note we here follow the train-test split from individual previous works.}
    \label{table:within}
\end{table}

\subsection{Cross-Dataset Evaluation}\label{sec:cross_dataset}

We first assess the generalization of our \methodname with the commonly used cross-dataset evaluation setting, where models are trained on one dataset and tested on an unseen dataset. 
Unlike previous studies that primarily report results for models trained on ETH-XGaze and/or Gaze360~\cite{cheng2022puregaze,xu2023learning,zhao2024improving,liu2021generalizing,liu2022jitter}, we evaluate models trained individually on five different gaze datasets and measure their cross-dataset performance on the remaining datasets. 
In \cref{table:cross_dataset}, each subtable corresponds to a training dataset, with each column representing a different test dataset. 
We compare several baseline models with our \methodname.
Note the 60 subjects from XGaze-Dense used during pre-training (\cref{sec:pretrain_data_prepare}) are excluded from the XGaze Test set.

Although, commonly, larger models tend to achieve better performance in typical computer vision tasks, the large ViT-based models (ViT-H, DINO-B, MoCo-v3-B) do not consistently surpass the ResNet-50 on gaze estimation as revealed in \cref{table:cross_dataset}.
This discrepancy indicates that simply increasing the model size without curated pre-training data is not effective for the downstream gaze estimation task.
Especially, the FaRL-B pre-trained on large face-centric data still cannot effectively handle gaze information across diverse datasets.

By contrast, the proposed \methodname-H, pre-trained on diverse facial datasets, demonstrates superior performance across nearly all settings. 
For completeness of fairness, we also show the \methodname-B based on the ViT-B backbone, which still outperforms similar model size models DINO-B and FaRL-B.
Note \methodname-B outperforms \methodname-H in some cases, which suggests again that our combination of large-model with our curated large-scale pre-training data effectively enhances ViT's ability to learn gaze-specific representations, rather than only increasing the size of the model. 
This answers our question that MAE pre-training with diverse facial data can strengthen the model's ability to capture fine-grained gaze geometry.

Besides, the results emphasize the importance of label range in training data.
While pre-training improves generalization across domains, it alone is insufficient when the training gaze data has a narrow label range.
For example, the model trained on MPIIFaceGaze (\cref{table:cross_m}) exhibits high errors on datasets such as XGaze Test (33.11 degrees) and Gaze360 (21.28 degrees), which have larger ranges of head poses and gaze directions than MPIIFaceGaze.
Thus, while MAE pre-training improves ViT's capacity to learn robust representations, diverse label coverage in the gaze estimation training stage remains crucial for achieving consistent performance in gaze estimation across domains.

\subsection{Comparison with Domain Generalization}\label{sec:domain_generalization}

\begin{table}[t]
    \centering
        \resizebox{0.88\linewidth}{!}{ 
        \setlength{\tabcolsep}{5.5pt}
            \begin{tabular}{l|ccccc}
            \toprule
            \textbf{Models \textbackslash~Test}  & \textbf{X\textsubscript{Test}} & \textbf{M} & \textbf{GC} & \textbf{E} & \textbf{G360} \\
            \midrule
            ResNet-50       & 16.31 & 4.94 & 6.71 & 7.89 & 18.69 \\
            GazeTR-50     & 15.47 & 4.94 & 7.22& 8.26 & 19.75\\
            ViT-L           & 19.32 & 5.09 & 6.33 & 8.98 & 22.68 \\
            \midrule
            FaRL-B          & 19.08 & 5.09 & 6.08 & 8.18 & 18.28 \\
            \methodname-B   & 11.78 & 4.73 & 5.86 & 6.31 & 12.41 \\
            \methodname-L   &\textbf{10.93}& 4.64 & 5.79 & 6.56 & 12.44  \\
            \methodname-H   & 11.29 &\textbf{4.51}& \textbf{5.47}&\textbf{5.88}&\textbf{12.37} \\
            \bottomrule
            \end{tabular}
        }
    \caption{
        Results of leave-one-dataset-out evaluation. 
        By taking the five gaze estimation datasets, we train the model on four datasets and test on the remaining one respectively. We show the results of each test dataset in the column. 
        Except for baselines, we evaluate three versions of \methodname with backbones of ViT-Base (\methodname-B), ViT-Large (\methodname-L), and ViT-Huge (\methodname-H).
    }
    \label{table:leave_out}
\end{table}

We further assemble the experiment results from our \methodname and compare them with the current SOTA domain generalization methods~\cite{cheng2022puregaze,xu2023learning,zhao2024improving,bao2024feature,yin2024clip}.
In \cref{table:sota}, we pick the typical cross-dataset setting, where the training dataset is either ETH-XGaze or Gaze360, and test on MPIIFaceGaze and screen targets (CS) subset of EYEDIAP.

Across all datasets, our \methodname-H model consistently achieves the lowest error, surpassing SOTA methods by significant margins.
Notably, while the ViT-H model pre-trained on ImageNet alone does not outperform domain generalization methods, our \methodname-H model pre-trained on large-scale data demonstrates significant improvements.
Again, this highlights that MAE pre-training on large-scale data greatly enhances ViT's ability for domain generalization, validating the effectiveness of our approach in improving gaze estimation across diverse domains.
Moreover, even \methodname-B, despite having a model size comparable to other SOTA methods, still achieves substantially better performance.

\begin{table}[t]
    \begin{center}
        \resizebox{0.95\linewidth}{!}{ 
        \setlength{\tabcolsep}{5.5pt}
            \begin{tabular}{l|ccccc}
            \toprule
            \textbf{Models \textbackslash~Test}  & \textbf{X\textsubscript{test}} & \textbf{M\textsubscript{test}} & \textbf{GC\textsubscript{test}} & \textbf{E\textsubscript{test}} & \textbf{G360\textsubscript{test}} \\
            \midrule
            ResNet-50   & 5.04 & 5.88 & 3.59 & 6.04 & 10.55 \\
            GazeTR-50 & 4.63 & 5.82 & 3.52 & 6.06 & 10.39 \\
            ViT-L & 5.24 & 5.42 & 3.66 & 6.30 & 10.54 \\
            \midrule
            \methodname-B & 4.75 & 5.40 & 3.37 & 5.52 & 9.64 \\
            \methodname-L & 4.67 & 5.12 & \textbf{3.17} & 5.33 & 9.29  \\
            \methodname-H & \textbf{4.46} & \textbf{5.08} & 3.20 & \textbf{5.16} & \textbf{9.07}  \\
            \bottomrule
            \end{tabular}
        }
    \end{center}
    \caption{
        Results of the joint-dataset evaluation protocol. We train the model on the gathered training splits of all five gaze estimation datasets.
        We test the model on the test set of each dataset individually, shown in each column.
    }
    \label{table:aggregate}
\end{table}

\begin{figure*}[t]
  \centering
  \begin{subfigure}{0.49\linewidth}
    \centering
    \includegraphics[width=0.88\linewidth]{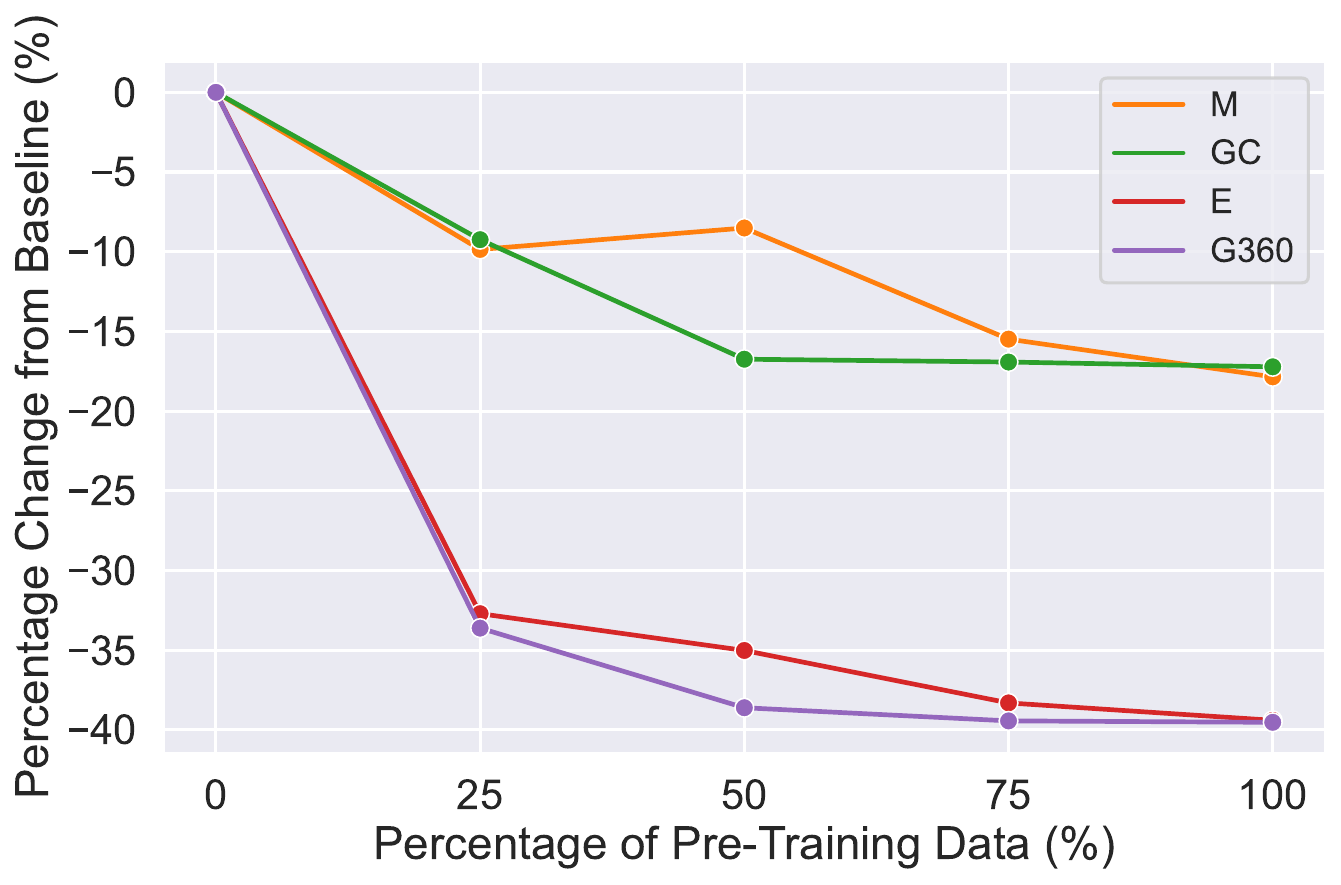}
    \caption{ The results when trained on ETH-XGaze.}
    \label{fig:plot_err_vs_data_size_xgaze}
  \end{subfigure}
  \hfill
  \begin{subfigure}{0.49\linewidth}
    \centering
    \includegraphics[width=0.9\linewidth]{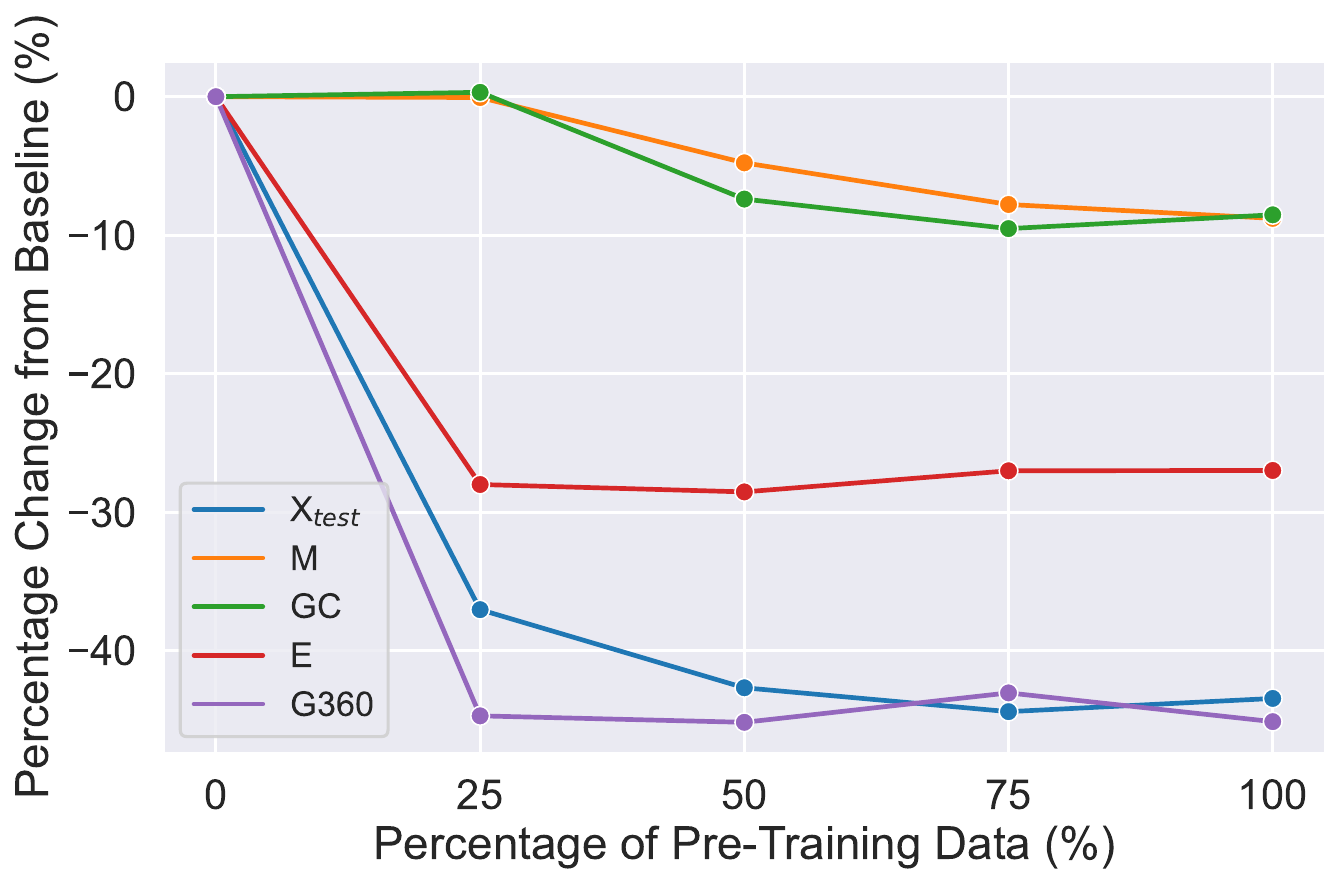}
    \caption{ The results of the leave-one-dataset-out evaluation.}
    \label{fig:plot_err_vs_data_size_leave_out}
  \end{subfigure}
  \caption{Effect of MAE pre-training data size on gaze estimation performance.
      The horizontal axis is the percentage of the pre-training data and the vertical axis is the percentage of the error reduction from the 0\% baseline.
    }
  \label{fig:plot_err_vs_data_size}
\end{figure*}

\subsection{Within-Dataset Evaluation}\label{sec:within}
To align with other SOTA methods, the splits for MPIIFaceGaze, ETH-XGaze, and EYEDIAP used in this section differ from the subject splits used elsewhere in the paper.
For GazeCapture and Gaze360, it is the same as defined in \cref{sec:exp_setting}.
For MPIIFaceGaze, we use the leave-one-subject-out protocol~\cite{abdelrahman2023l2cs,shi2024agent,ververas20253dgazenet}. 
For ETH-XGaze, we follow the train-test split provided in~\cite{ververas20253dgazenet}.
For EYEDIAP, we apply the four-fold validation scheme from~\cite{cheng2022gaze,cheng2024appearance}. 
The results in \cref{table:within} demonstrate that \methodname-H outperforms SOTA on most datasets, except MPIIFaceGaze, indicating that the proposed pre-training enhances generalizability even within the same dataset.

\subsection{Leave-One-Dataset-Out Evaluation}\label{sec:leave_one_out}

We conduct a \textit{leave-one-dataset-out} evaluation to further assess the generalizability of \methodname.
Using five gaze estimation datasets, we train the model on a combination of four datasets and test it on the held-out dataset. 
In this manner, we assess the upper-bound performance that can be achieved in each domain by maximizing the use of existing datasets.
We compare the baseline models of ResNet-50, GazeTR-50, and ViT-H that are pre-trained on ImageNet.
We present the three versions of \methodname with different backbones of ViT-Base, ViT-Large, and ViT-Huge.

We show the results in \cref{table:leave_out} with each column showing the results on each test dataset. 
Across test datasets, the proposed \methodname with three backbone sizes consistently surpasses all baselines, achieving substantial error reductions. 
By comparing the \methodname with three sizes of backbones, we can see that the trend clearly shows the larger model achieves better performances in general.
Notably, large-margin improvements from \methodname happen in the XGaze and Gaze360 test sets, highlighting that \methodname significantly enhances ViT’s generalization ability across diverse domains by fostering robust representations of facial appearance and image quality.

\subsection{Joint-Dataset Evaluation}\label{sec:joint}

Building upon the leave-one-dataset-out evaluation, we conduct a new evaluation protocol \textit{joint-dataset} evaluation.
In this setting, we gather the training sets of all five datasets, including XGaze Train, MPIIFaceGaze Train, GazeCapture Train, EYEDIAP Train, and Gaze360 Train to form a large and comprehensive training set.
The model trained on this large training set is expected to be the most powerful gaze estimator that we can acquire.
We test the trained model on the test sets of each dataset.
It has another meaning of generalization, in which one model performs gaze estimation across multiple data domains.

\Cref{table:aggregate} shows the evaluation results where each column represents a specific test dataset.
As expected, our \methodname-H model improves over other methods, achieving the lowest error on most test datasets compared to all baselines and smaller backbone versions.
Since the model is trained on data that covers the domains of all test datasets, the resulting errors are generally lower and saturated compared to other settings. 
It aligns with our purpose of training one model for all domains with the lowest error.

\subsection{Ablation Studies}\label{sec:ablation}

\subsubsection{Effect of Pre-Training Data Size}
The large-scale pre-training with MAE is the most critical factor for our \methodname. 
To analyze the impact of MAE pre-training data size, we vary the amount of data used for the pre-training stage. 
With the full pre-training data of around \SI{1.6}{\mega{}} images, we randomly sample subsets from each component dataset (\cref{sec:pretrain_data_prepare}) to create 25\%, 50\%, and 75\% subsets of the full data.
We use the \methodname-L as the backbone, and the 0\% refers to the ImageNet pre-trained ViT-L.

We present two experiment settings in \cref{fig:plot_err_vs_data_size}, the XGaze training (left) and the leave-one-dataset-out setting (right).
Beginning with the 0\% baseline, we illustrate the percentage reduction in error (vertical axis) as the pre-training dataset size (horizontal axis) increases, allowing us to capture and compare relative performance improvements across various pre-training levels.

Overall, the results indicate that the pre-trained model on a larger subset consistently achieves lower error across all test domains.
There are notable performance improvements with 25\% and 50\% of the data, and the improvement gap becomes smaller when reaching the 75\% subset to the full data, which is expected due to a sufficient amount of data diversity for the training.
These results confirm that increasing the amount of data in MAE pre-training strengthens ViT’s representation learning, leading to improved accuracy and generalization across diverse gaze estimation tasks.

\begin{table}[t]
    \centering
    \resizebox{1.0\linewidth}{!}{ 
    \setlength{\tabcolsep}{4.5pt}
         \begin{tabular}{l|ccccc}
        \toprule
        \textbf{Models \textbackslash~Test} & \textbf{X\textsubscript{test}} & \textbf{M} & \textbf{GC} & \textbf{E} & \textbf{G360} \\
        \midrule
        FaRL-B~\cite{zheng2022general}  & 19.08 & 5.09 & 6.08 & 8.18 & 18.28 \\
        \midrule
        \methodname-B (\textit{Real, {\small limited poses}}) & 16.70 & 5.29 & 6.87 & 6.57 & 13.78 \\
        \methodname-B (\textit{Real, {\small w/o norm.}}) & 14.56 & 4.95 & 6.27 & 6.93 & 14.48 \\
        \methodname-B (\textit{Real}) & 11.95 & 4.86 & 6.14 & \textbf{6.26} & 12.71 \\
        \methodname-B & \textbf{11.78} & \textbf{4.73} & \textbf{5.86} & 6.31 & \textbf{12.41} \\
        \bottomrule
        \end{tabular}
        }
    \caption{
    Comparison of different formats of the input data in the leave-one-dataset-out evaluation setting. Each column shows the results on each test dataset. For fair compassion with FaRL-B~\cite{zheng2022general}, we use the ViT-B backbone (\methodname-B). 
    \textit{Real} stands for the combination of our real datasets CelebV-Text, VFHQ, and VGGFace2.
    }
\label{table:ablation_data_format}
\end{table}

\subsubsection{Effect of Pre-Training Data Attributes}
To better understand the performance gap from FaRL-B~\cite{zheng2022general}, we examine the impact of different data in \cref{table:ablation_data_format}.
To assess the effect of the synthetic data, we first limit the pre-training dataset to the real datasets: CelebV-Text, VFHQ, and VGGFace2 (\textit{Real}).
To evaluate the impact of pre-processing specifically for gaze estimation, we also assess the performance when pre-training is conducted using the loose face bounding boxes directly, without applying any data normalization~\cite{zhang2018revisiting} (\textit{Real, w/o norm.}).
To evaluate the effect of wide head pose range, we limit head pose variability by filtering out samples with a pitch-yaw $L_2$ norm exceeding 10 degrees, reducing the dataset size to about 20\% (\textit{Real, limited poses}). 
Note that the different amounts of data could also affect the model's performance.
We use the leave-one-dataset-out evaluation protocol, given the trade-off between the task difficulty and simplification.

The results demonstrate that the \methodname-B trained on full data performs best compared to other baselines in almost all settings.
Models without normalization (\textit{Real, w/o norm.}) show substantially degraded performance despite using identical data.
Models with limited pose ranges (\textit{Real, limited poses}) also perform significantly worse than those trained on a more diverse range.
Integrating real and synthetic data yields the best results, which can be attributed to combining naturalistic appearance variations from real data with controlled pose variations from synthetic data.

\section{Conclusion}
\label{sec:conclusion}
In this paper, we provide the first extensive study of the self-supervised large-scale pre-training on gaze estimation. 
Through our extensive experimentation, we have established key principles for effective pre-training in gaze estimation.
With careful data curation for the MAE pre-training, the proposed \methodname achieves distinguished performances for cross-dataset, leave-one-dataset-out, and joint-dataset evaluations compared to current SOTAs.
Interestingly, we show the importance of the pre-training data selection and pre-processing for the final performance, rather than simply gathering a large amount of data.
Looking forward, we believe there remains significant potential for further refinement of pre-training data selection based on the principles identified in this work. 
Another potential improvement option could be using a large-size face image input for the high-resolution scenarios.

{
    \small
    \bibliographystyle{ieeenat_fullname}
    \bibliography{main}
}

\clearpage
\setcounter{page}{1}
\setcounter{figure}{0}
\setcounter{table}{0}
\setcounter{equation}{0}

\appendix

\begin{center}
    {\Large \bfseries Supplementary Materials}\\[1em]
\end{center}

\begin{table*}[t]
    \centering
    \resizebox{1.0\linewidth}{!}{ 
    \setlength{\tabcolsep}{4.5pt}
         \begin{tabular}{l|ccccc}
        \toprule
        \textbf{Training Data \textbackslash~Test}  & \textbf{X\textsubscript{test}} & \textbf{M\textsubscript{test}} & \textbf{GC\textsubscript{test}} & \textbf{E\textsubscript{test}} & \textbf{G360\textsubscript{test}} \\
        \midrule
        \rowcolor{LightGray} \multicolumn{6}{l}{ResNet-50}  \\
         \textit{same-domain}       & 5.25  & 5.11   & 3.49  &  8.51 & 11.87\\
         \textit{leave-one-dataset-out}    & 16.31 (\increase{210.7}) & 6.23 (\increase{21.9}) & 6.35 (\increase{82.0}) & 8.25 (\reduce{3.1}) & 20.38 (\increase{71.7}) \\
         \textit{joint-dataset} & \textbf{5.04} (\reduce{4.0}) & 5.88 (\increase{15.1}) & 3.59 (\increase{2.9}) & \textbf{6.04} (\reduce{29.0}) & \textbf{10.55} (\reduce{11.1}) \\
         \midrule
        \rowcolor{LightGray}
        \multicolumn{6}{l}{\methodname-H} \\
        \textit{same-domain}   & 4.62 & 5.19 & 3.01 & 6.11 & 9.44  \\
         \textit{leave-one-dataset-out} & 11.29 (\increase{144.4}) & 5.22 (\increase{0.6}) & 5.13 (\increase{70.4}) & 6.14 (\increase{0.5}) & 13.12 (\increase{39.0}) \\
          \textit{joint-dataset} & \textbf{4.46} (\reduce{3.5}) & \textbf{5.08} (\reduce{2.1}) & 3.20 (\increase{6.3}) & \textbf{5.16} (\reduce{15.6}) & \textbf{9.07} (\reduce{3.9})  \\
        \bottomrule
        \end{tabular}
        }
    \caption{
        Comparison of different training data configurations for gaze estimation.
        Each column represents a specific test dataset: XGaze Test, MPIIFaceGaze Test, GazeCapture Test, EYEDIAP Test, and Gaze360 Test.
        Each row corresponds to a training configuration:
        \textit{Same-domain} means training on the same domain as the test set,
        \textit{leave-one-dataset-out} means training on the remaining four datasets other than the test set, and \textit{joint-dataset} means training on the aggregated Train split of all five datasets.
        The percentages in parentheses indicate the reduction or increment compared to the \textit{same-domain} results, where lower errors indicate better performance.
        For the \textit{leave-one-dataset-out} configuration, the errors reported here are on the Test splits, while the main paper reports errors on the entire dataset.
    }
\label{table:comprehensive}
\end{table*}

\noindent In this supplementary material, we first provide an analysis of the effect of combining multiple domains.
Then, we include additional ablations to investigate the effects of color-jitter augmentation and pixel normalization during the pre-training.
Finally, we present qualitative results, highlighting images captured under diverse and challenging conditions.

\section{Analysis on Combining Multiple Domains}\label{sec:supp_data}

We analyze the effect of different training data configurations on gaze estimation performance.
Specifically, we compare three configurations: training on the same domain (\textit{same-domain}), training on multiple domains excluding the testing domain (\textit{leave-one-dataset-out}), and training on multiple domains including the testing domain (\textit{joint-dataset}).

\Cref{table:comprehensive} shows the comparison of gaze errors for these configurations.
Each column corresponds to a specific test dataset: XGaze Test, MPIIFaceGaze Test, GazeCapture Test, EYEDIAP Test, and Gaze360 Test, while each row represents a training configuration.
This \textit{same-domain} setting is different from the \textit{within-dataset} in the main paper. 
We use the splits defined in Sec.~4.1 of the main paper.
Especially, please note that for MPIIFaceGaze dataset, we train the model on the first 10 subjects and test on the remaining five subjects, different from the typical leave-one-subject-out protocol~\cite{abdelrahman2023l2cs,shi2024agent,ververas20253dgazenet}.

The percentages in parentheses indicate the reduction or increment compared to the \textit{same-domain} results, where lower errors indicate better performance.
Note that, for the \textit{leave-one-dataset-out} configuration, errors on the entire left-out dataset are reported in the main paper, but here we present errors on the Test split to align with the other configurations that require dataset splits.

\paragraph{\textit{Same-domain}}
In general, training and testing on the same domain (\textit{same-domain}) yields the best results, even though datasets combined from multiple domains have the potential to be more diverse.
This emphasizes the persistent challenge of achieving optimal performance when using data from different domains.
The exception observed for E\textsubscript{test} with the ResNet-50 backbone may be attributed to the limited number of samples in the EYEDIAP Train split.

\paragraph{\textit{Leave-one-dataset-out}}
In the \textit{leave-one-dataset-out} configuration, we observe varying tendencies across different test datasets. 
Some datasets achieve errors comparable to the \textit{same-domain} results, while others remain challenging. 
For instance, for M\textsubscript{test} and E\textsubscript{test}, which are relatively less complex, the remaining four datasets provide sufficient information to achieve good performance.
In contrast, for X\textsubscript{test}, GC\textsubscript{test}, and G360\textsubscript{test}, the remaining four datasets fail to fully capture the critical factors required for optimal performance.
This variation highlights the strong dependence of performance on the attributes of the training data.

Importantly, our \methodname-H demonstrates a smaller performance gap compared to ResNet-50 in most cases, with the only exception being EYEDIAP, where the difference is marginal. 
This suggests that \methodname-H is better equipped to learn gaze representations from out-of-domain data with less overfitting, underscoring its enhanced generalization capability.

\paragraph{\textit{Joint-dataset}}

Overall, the \textit{joint-dataset} configuration demonstrates significant promise, creating a single model robust across multiple test domains. 
For \methodname-H, the only exception is GC\textsubscript{test}, where the \textit{joint-dataset} configuration produces a slightly higher error (3.01$\rightarrow$3.20). 
Although this suggests some negative effects from the other four datasets, the effects remain marginal.
While the improvement percentages for \methodname-H are smaller compared to ResNet-50, the absolute errors are consistently lower.

\begin{table}[t]
    \centering
    \resizebox{0.8\linewidth}{!}{ 
    \centering
    \setlength{\tabcolsep}{6pt}
        \begin{tabular}{ccc|cccc}
        \toprule
        \textbf{\textit{Real}} & \textbf{\textit{Syn.}} & \textbf{\textit{NV.}}  & \textbf{M} & \textbf{GC} & \textbf{E} & \textbf{G360} \\
        \midrule
        \checkmark &            &            & 6.79 & 7.81 & 6.86 & 12.93 \\
        \checkmark & \checkmark &            & 6.57 & 7.37 & \textbf{6.51} & 13.23 \\
        \checkmark & \checkmark & \checkmark & \textbf{6.21} & \textbf{7.35} & 6.64 & \textbf{12.18} \\
        \bottomrule
        \end{tabular}
    }
    \caption{
    We ablate the pre-training facial datasets by comparing real, synthetic, and novel-rendered images.
    The comparison is performed on the \methodname-B network, followed by training on XGaze.
    The last row represents the full-dataset setting.
    }
\label{table:ablation_data_type}
\end{table}

\begin{figure*}[t]
  \centering
  \begin{subfigure}{0.99\linewidth}
    \centering
    \includegraphics[width=0.99\linewidth]{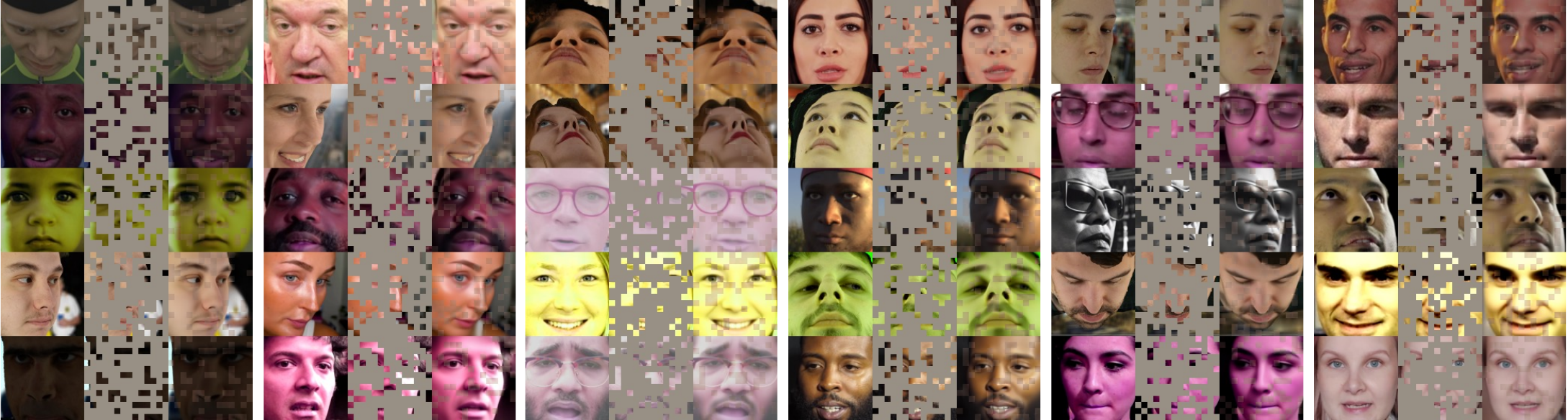}
    \caption{ MAE reconstruction examples without pixel normalization.}
    \label{fig:no_pixel_norm}
  \end{subfigure}
  \begin{subfigure}{0.99\linewidth}
    \centering
    \includegraphics[width=0.99\linewidth]{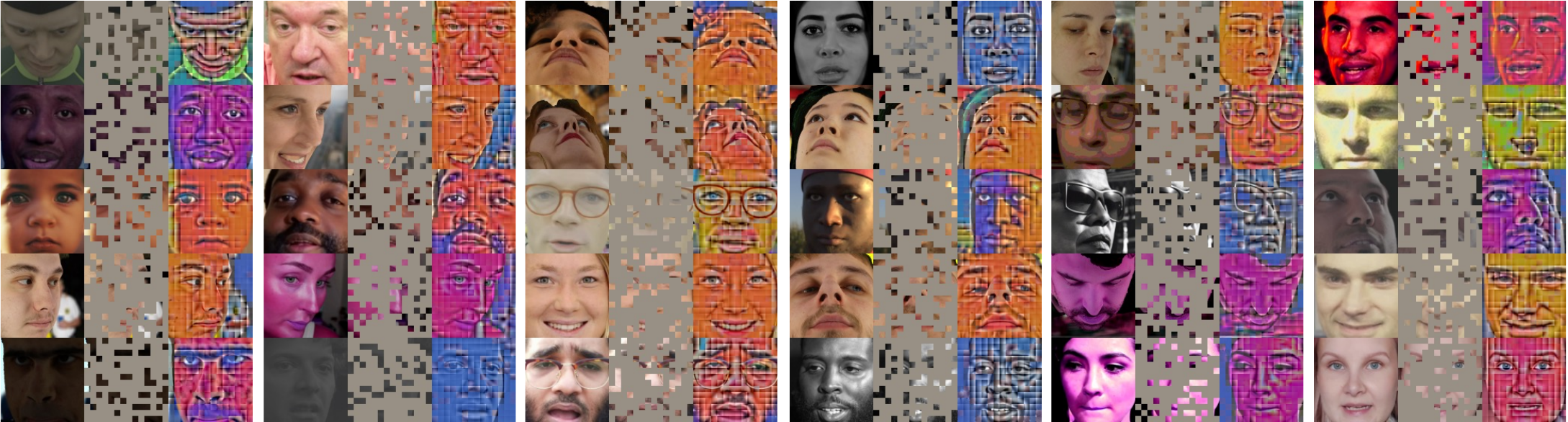}
    \caption{ MAE reconstruction examples with pixel normalization (Proposed).}
    \label{fig:w_pixel_norm}
  \end{subfigure}
  \caption{Examples comparison of the pixel normalization during the MAE pre-training.
  The left, middle, and right columns show the original image, masked input, and the reconstructed image, respectively.
    }
  \label{fig:pixel_norm_samples}
\end{figure*}

\begin{figure}[t]
  \centering
    \includegraphics[width=0.9\linewidth]{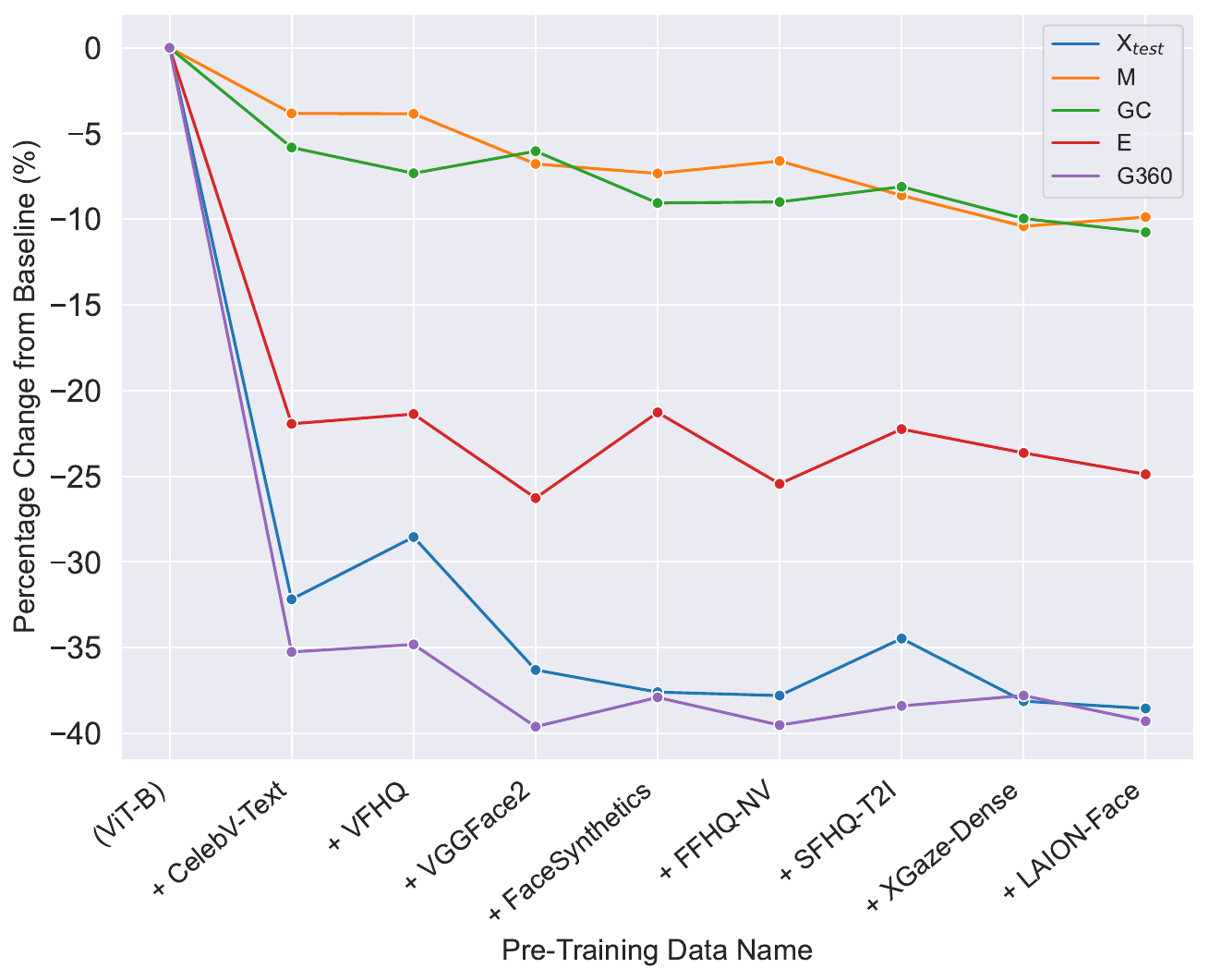}
  \caption{
        Effect of MAE pre-training dataset composition on downstream gaze estimation performance.  
        The horizontal axis represents the incremental accumulation of datasets, while the vertical axis shows the percentage reduction in error relative to the first CelebV-Text dataset~\cite{yu2023celebv}.
    }
  \label{fig:plot_err_vs_data_composition}
\end{figure}

\section{Additional Ablation Studies on Pre-Training}

\paragraph{Effect of Pre-Training Dataset Composition}

Beyond the overall pre-training dataset size, the composition of the dataset also plays a critical role in learning effective gaze representations.
To investigate the impact of different facial dataset components, we conduct an experiment where we incrementally accumulate datasets during the MAE pre-training stage and analyze their effect on the downstream gaze estimation performance.
Starting with CelebV-Text~\cite{yu2023celebv}, we progressively add datasets for pre-training and evaluate model separately on gaze estimation.
Each pre-trained model is subsequently trained on gaze datasets using the same \textit{leave-one-dataset-out} protocol.
\Cref{fig:plot_err_vs_data_composition} illustrates the error change across different test sets as more datasets are included in pre-training.

Overall, the results indicate that adding more diverse data during pre-training generally enhances gaze generalization.
However, there are exceptions that adding a model can result in increased error for specific test sets.
For example, adding VFHQ increases the error on the XGaze Test set from 12.65 to 13.32, while including SFHQ-T2I causes performance fluctuations across different benchmarks.
This suggests that certain dataset attributes may not align well with particular test distributions, leading to suboptimal transferability.
On the other hand, datasets such as VGGFace2 and XGaze-Dense provide performance improvements on most test sets.
Additionally, performance gains becomes marginal as the dataset number increases, aligning with the analysis of pre-training data size in main paper.

In conclusion, dataset diversity plays a crucial role in improving MAE pre-training for gaze estimation.
A more detailed analysis of dataset attributes and their impact on gaze estimation remains an open research question, which we leave for future work.
Nonetheless, our empirical results suggest that increasing data diversity in pre-training tends to improve model performance across various test domains.

\paragraph{Effect of Novel-View Synthesis Data in Pre-Training}\label{sec:ablation_novel_view_data}
To examine the effect of novel-view synthesis in pre-training data, we conduct further experiments separating these two elements.
In \cref{table:ablation_data_type}, we conduct an ablation study by varying data subsets during the pre-training: real datasets (CelebV-Text, VGGFace2, and VFHQ), synthetic datasets (FaceSynthetics and SFHQ-T2I), and novel-view-rendered datasets (FFHQ-NV and XGaze-Dense).
We use the \methodname-B to conduct the experiment due to its time efficiency.
After pre-training, we train on XGaze and test on the rest of the four datasets. 

The results further clarify the effect of different data types on the model's generalizability.
Adding synthetic data (\textit{Real + Syn.}) reduces errors in several test domains compared to using only real data, suggesting the variability of the synthetic data contributes to generalization.
Further incorporating novel-view data (\textit{Real + Syn. + NV}) provides additional performance gains, especially in head-pose generalization, likely due to the expanded range of facial orientations.
This finding supports the idea that a mix of real, synthetic, and novel-view data in MAE pre-training strengthens ViT’s representation learning.

\paragraph{Effect of Pixel Normalization}\label{sec:ablation_pixel_norm}
The patch normalization technique is applied during the MAE pre-training as suggested in~\cite{he2022masked} which is different from reconstructing the natural image, as shown in \cref{fig:pixel_norm_samples}.
We compare models pre-trained with and without patch normalization to investigate its impact.

\paragraph{Effect of Color-Jitter Augmentation}\label{sec:ablation_augmentation}
Color jittering introduces randomness in brightness, contrast, saturation, and hue to simulate diverse lighting conditions, enhancing the robustness of learned features.
We compare models pre-trained with and without color-jitter augmentation to investigate its impact.

\paragraph{Results}
We use the \methodname-B model as the backbone and compare different pre-training settings, followed by training on XGaze and testing on the remaining four datasets.
\Cref{table:ablation_colorjitter_pixelnorm} demonstrates that both color-jitter augmentation and pixel normalization contribute to improved gaze estimation performance, highlighting their benefits for the generalization of the pre-trained model.
Notably, pixel normalization consistently improves performance across all test datasets, aligning with the observations in the original MAE paper~\cite{he2022masked}, which showed that pixel normalization enhances representation learning.

\begin{table}[t]
    \centering
    \resizebox{0.97\linewidth}{!}{ 
    \centering
    \setlength{\tabcolsep}{6pt}
        \begin{tabular}{cc|cccc}
        \toprule
        \textbf{Color-Jitter} & \textbf{Pixel Norm.} & \textbf{M} & \textbf{GC} & \textbf{E} & \textbf{G360} \\
        \midrule
         \XSolidBrush & \XSolidBrush     & 7.52 & 8.01 &  8.56 & 14.14 \\
        \checkmark & \XSolidBrush   &  7.17 & 8.23 & 8.03 & 14.03 \\
        \XSolidBrush   & \checkmark &  7.18 & 7.94 & 8.05 & 13.66 \\
        \checkmark & \checkmark & \textbf{6.21} & \textbf{7.35} & \textbf{6.64} & \textbf{12.18} \\
        \bottomrule
        \end{tabular}
    }
    \caption{
        Ablation studies on the pre-training, comparing the effect of the color-jitter augmentation and the pixel normalization.
        During the gaze estimation training, we train the model using XGaze and test on the other four datasets to evaluate the generalizability.
    }
\label{table:ablation_colorjitter_pixelnorm}
\end{table}

\section{Comparison with the SOTAs}
3DGazeNet~\cite{ververas20253dgazenet} collects in-the-wild face images with pseudo gaze labels and applies multi-view synthesis to obtain an augmented dataset ITWG-MV.
To account for the difference in test data settings, we compare 3DGazeNet with \methodname-H separately in \cref{table:supp_sota_3dgn}.
The results demonstrate that \methodname-H outperforms 3DGazeNet in all domain generalization settings.

\paragraph{Re-implementation}
In the main paper, we compared our \methodname-H model with state-of-the-art (SOTA) methods using their reported results. 
It is important to note that minor discrepancies may arise due to differences in our data pre-processing compared to prior work~\cite{cheng2022puregaze,xu2023learning,zhao2024improving}. 
To ensure a fair comparison, we re-implemented ResNet-18 and PureGaze~\cite{cheng2022puregaze} using our pre-processed datasets, aligning them with the reported results~\cite{cheng2022puregaze,zhao2024improving}. 
The re-implementation results, alongside the reported values, are summarized in \cref{table:supp_sota}.

While minor differences exist between our re-implementation and the reported values, the improvements achieved by our \methodname-H model remain significant, demonstrating its superior performance across all domain generalization tasks.

\begin{table}[t]
\begin{center}
    \resizebox{0.99\linewidth}{!}{ 
    \setlength{\tabcolsep}{4pt}
        \begin{tabular}{l|cccc}
        \toprule
        \textbf{Models}  & \textbf{X}$\rightarrow$\textbf{M} & \textbf{X}$\rightarrow$\textbf{GC}& \textbf{G360}$\rightarrow$\textbf{M} & \textbf{G360}$\rightarrow$\textbf{GC} \\
        \midrule
        \rowcolor{LightGray}
        3DGazeNet$^{\dagger}$~\cite{ververas20253dgazenet} & 6.0 & 7.8 & 6.3 & 8.0  \\
        \midrule
        \methodname-H & \textbf{5.57} &  \textbf{6.56} & \textbf{5.43} & \textbf{6.48} \\
        \bottomrule
        \end{tabular}
}
\end{center}
\caption{
    Domain generalization compared with SOTA methods. 
    The results marked with $^{\dagger}$ are directly cited from previous studies~\cite{ververas20253dgazenet}.
}
\label{table:supp_sota_3dgn}
\end{table}

\begin{table}[t]
\begin{center}
    \resizebox{0.99\linewidth}{!}{ 
    \setlength{\tabcolsep}{4pt}
        \begin{tabular}{l|cccc}
        \toprule
        \textbf{Models}  & \textbf{X}$\rightarrow$\textbf{M} & \textbf{X}$\rightarrow$\textbf{E\textsubscript{CS}}& \textbf{G360}$\rightarrow$\textbf{M} & \textbf{G360}$\rightarrow$\textbf{E\textsubscript{CS}} \\
        \midrule
        ResNet-18   & 7.57 & 9.54 & 9.24  & 8.07  \\
        \rowcolor{LightGray}
        ResNet-18$^{\dagger}$~\cite{zhao2024improving}  & 8.02 & 9.11 & 8.04 & 9.20 \\
        PureGaze & 6.68 & 7.62 & 8.87 & 10.53 \\ 
        \rowcolor{LightGray}
        PureGaze$^{\dagger}$~\cite{cheng2022puregaze}  & 7.08 & 7.48 & 9.28 & 9.32 \\
        \midrule
        \methodname-H & \textbf{5.57} &  \textbf{4.65} & \textbf{5.43} & \textbf{5.35} \\
        \bottomrule
        \end{tabular}

}
\end{center}
\caption{
    Domain generalization compared with SOTA methods and their re-implementations. 
    The results marked with $^{\dagger}$ are cited from previous studies~\cite{cheng2022puregaze,zhao2024improving}, and the rest of the results are based on our implementation.
}
\label{table:supp_sota}
\end{table}

\section{Implementation Details}

\paragraph{Novel-Rendered Data Preparation}

To render images from novel views, we follow the rendering approach described in~\cite{qin2022learning}.
To control the head pose, we randomly generate target head poses and compute the corresponding rotation matrices to apply to the 3D face models. 
During the rendering process, 40\% of the images are assigned a random background color, while the remaining 60\% use random scene images from the Places365 dataset~\cite{zhou2017places} as background. 
Additionally, to simulate varied lighting conditions, half of the rendered images are adjusted to have lower ambient light intensity, ranging from $0.2$ to $0.75$.

All face images in our method are in the size of \num{224} $\times$ \num{224} after the data normalization process~\cite{zhang2018revisiting}.
When the camera parameters are unknown, we use a camera matrix with focal length $f$ set to the image width and principal point $(c_x, c_y)$ set to half the image height and width.

\paragraph{Pre-Training}
We apply random color jitter augmentation with a probability of 0.5 and the following parameters: hue in the range $[-0.15, 0.15]$, saturation in $[0.8, 1.2]$, contrast in $[0.4, 1.8]$, and brightness in $[0.7, 1.3]$.
We apply random grayscale with a probability of 0.05 on all images.

\paragraph{Gaze Estimation Training}

We use the Adam optimizer~\cite{kingma2014adam} with a learning rate of \num{1e-4} and a weight decay of \num{1e-6} for all experiments.
For experiments with ResNet-50 and GazeTR-50, we set the batch size to 128 and decay the learning rate by 0.1 every five epochs, with a total of 12 epochs.
For cross-dataset evaluation with \methodname-H, we use a batch size of 128 and train the model for eight epochs with the one-cycle learning rate schedule~\cite{smith2019super}.
For \textit{leave-one-dataset-out} and \textit{joint-dataset} evaluations, we set the batch size to 160 with 12 epochs.

\section{Qualitative Results}
In this section, we present additional qualitative results using the \methodname-H model trained on the aggregated datasets under the \textit{joint-dataset} setting. 
We employ an off-the-shelf facial landmark detector~\cite{bulat2017far} to extract landmarks and perform data normalization.
Gaze estimation is conducted on the normalized images, and the results are de-normalized back to the original image for visualization.
For reference, we also include the normalized faces alongside the original images.

\begin{figure*}[t]
  \centering
   \includegraphics[width=0.97\linewidth]{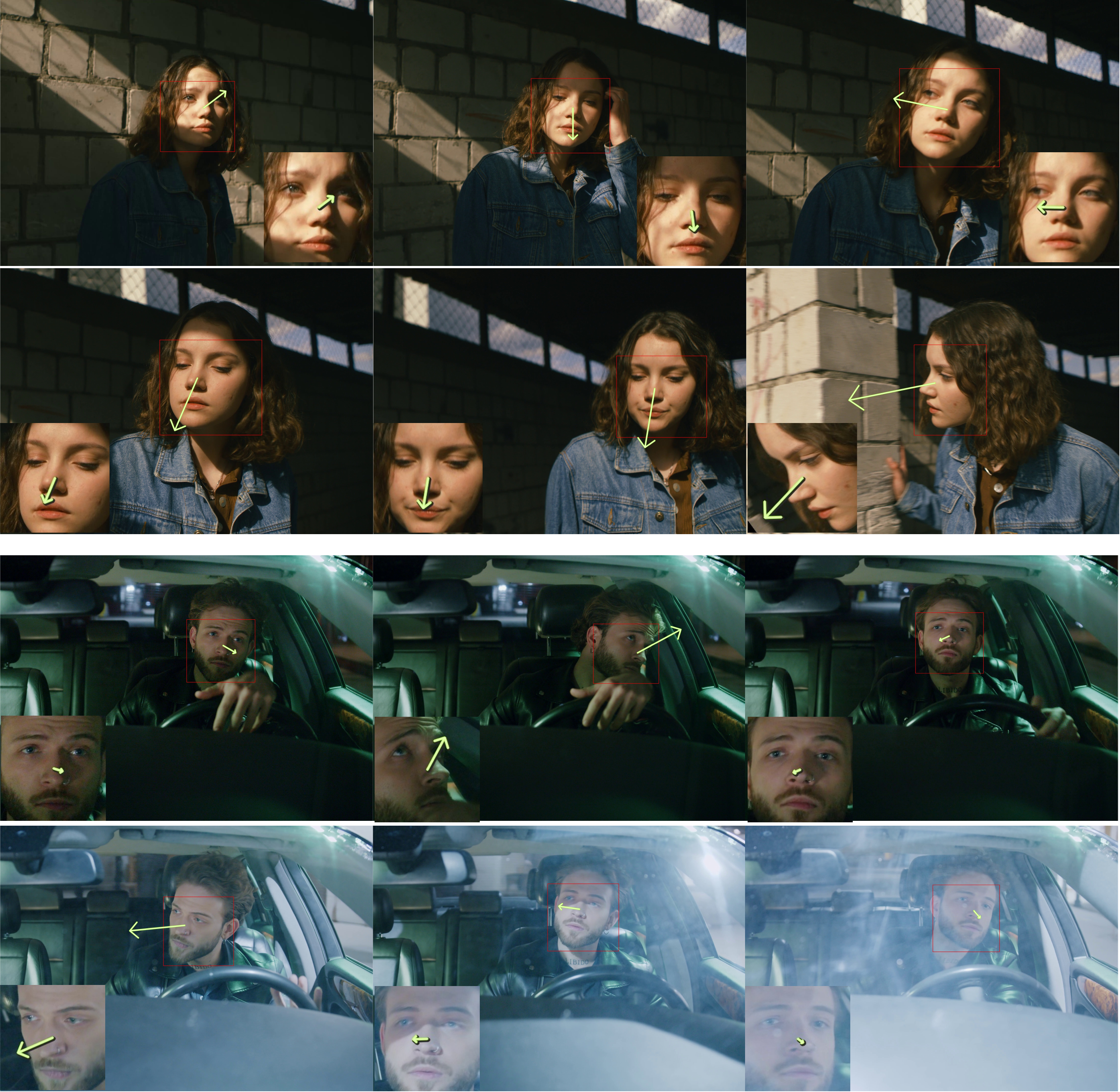}
   \caption{Qualitative results from various in-the-wild video examples.
   The normalized input images are displayed alongside the original image for reference.
   }
   \label{fig:results_qual_wild_one_person}
\end{figure*}

\begin{figure*}[t]
  \centering
  \begin{subfigure}{0.99\linewidth}
    \centering
    \includegraphics[width=0.97\linewidth]{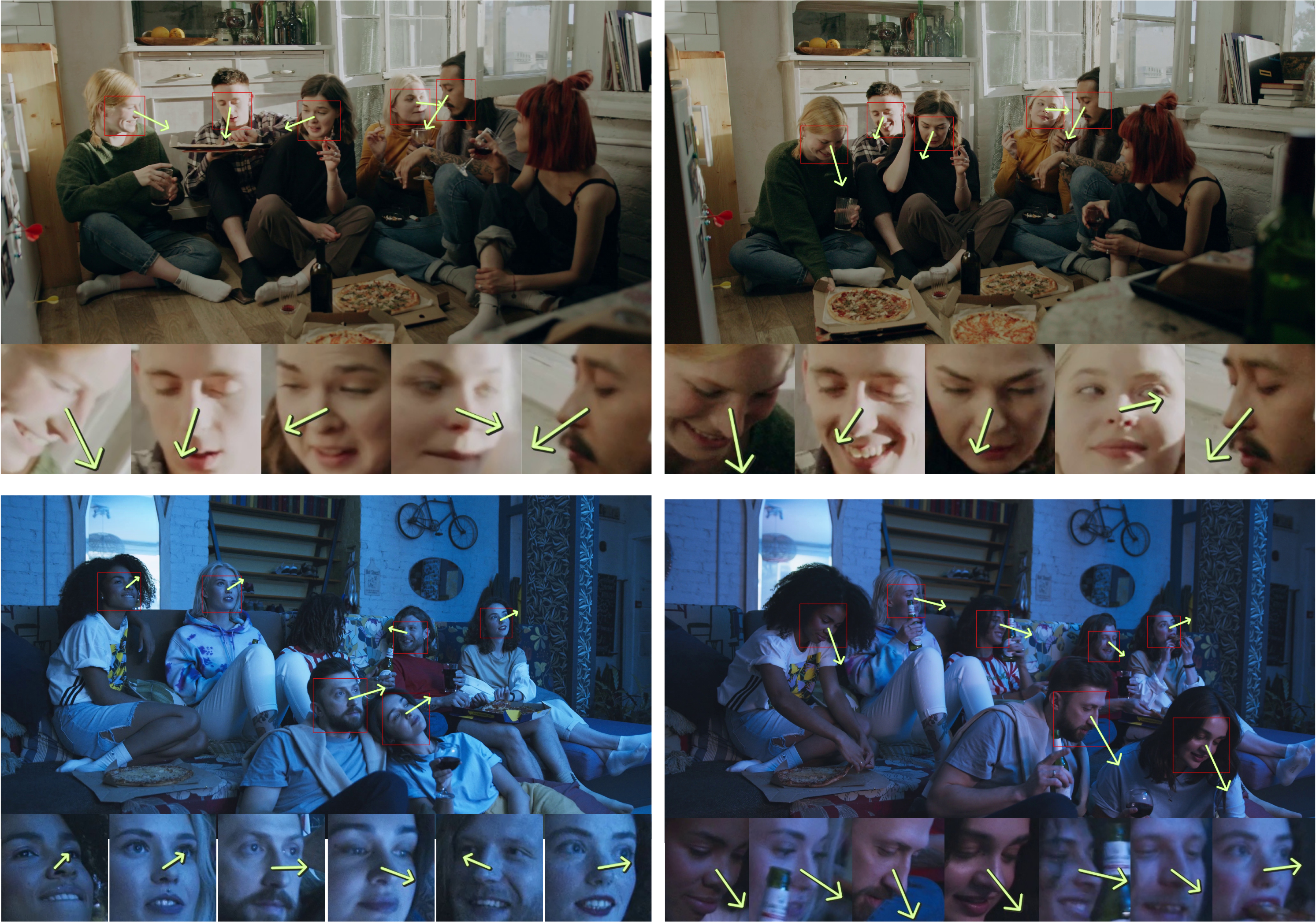}
    \caption{ Qualitative results of in-the-wild video examples from public domain.
    }
    \label{fig:results_qual_wild_public}
  \end{subfigure}
  \begin{subfigure}{0.99\linewidth}
    \centering
    \includegraphics[width=0.97\linewidth]{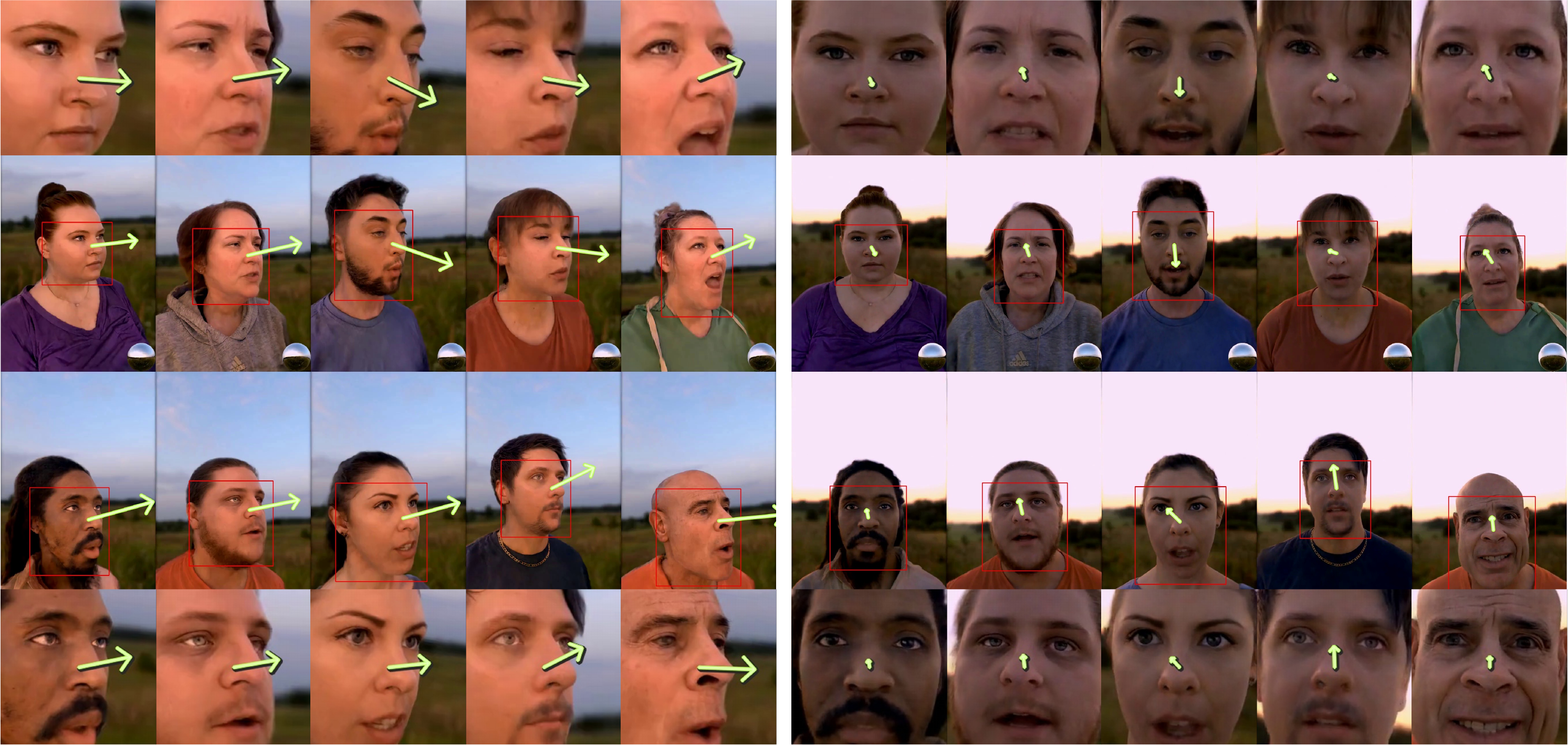}
    \caption{ Qualitative results of synthetic faces from URAvatar~\cite{li2024uravatar} .}
    \label{fig:results_qual_uravatar}
  \end{subfigure}
      \caption{Qualitative results of in-the-wild video and synthetic video.
      The normalized input images are displayed alongside the original image for reference.
    }
  \label{fig:results_qual_wild}
\end{figure*}

\begin{figure*}[t]
  \centering
   \includegraphics[width=0.97\linewidth]{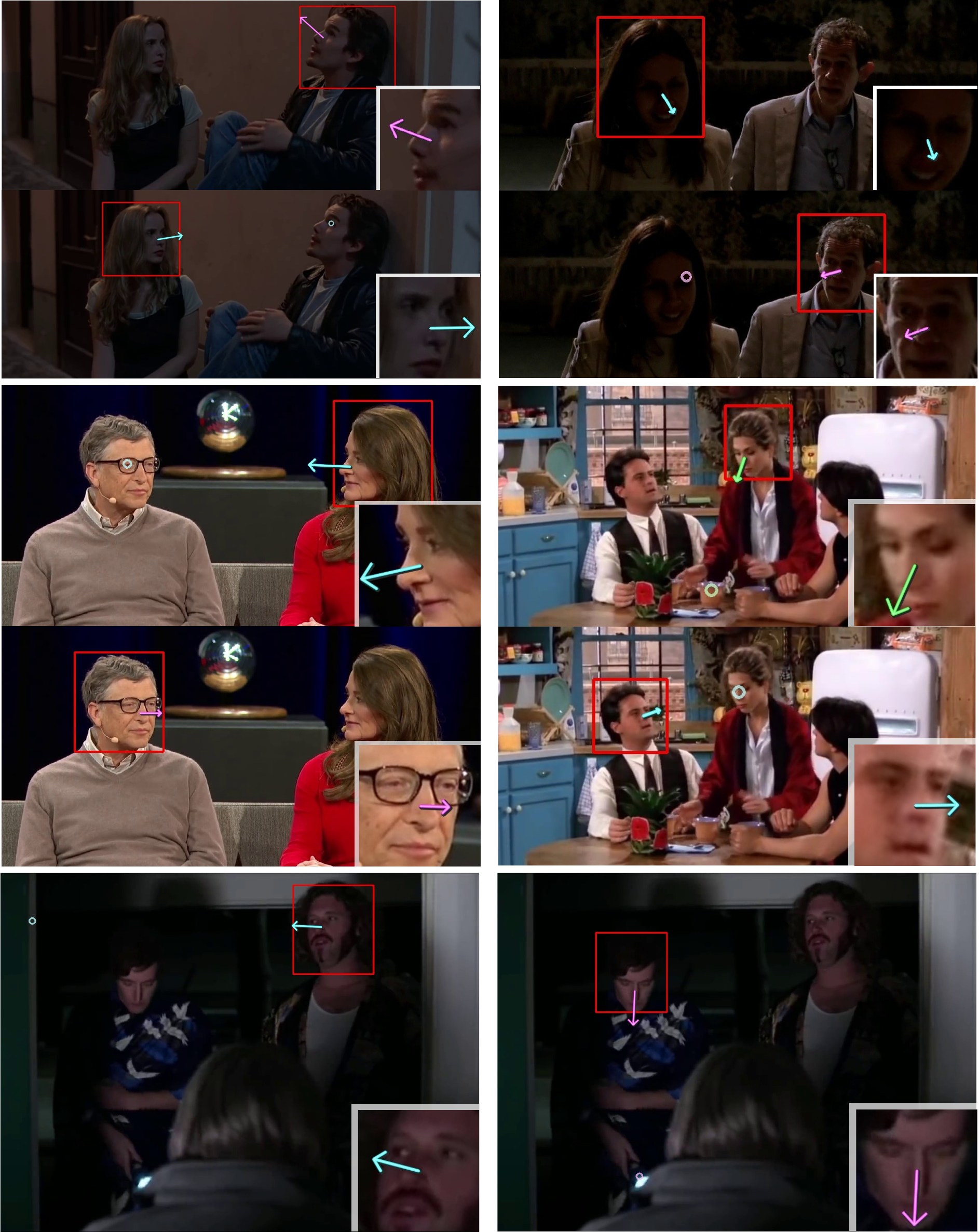}
   \caption{Qualitative results of examples from the VideoAttentionTarget dataset~\cite{chong2020detecting}.
   Gaze targets are visualized when annotated within the image frame, as some targets may be out of frame.
   The normalized input images are displayed alongside the original image for reference.
   }
   \label{fig:results_qual_VAT}
\end{figure*}

\Cref{fig:results_qual_wild_one_person} and \cref{fig:results_qual_wild} showcase examples from various in-the-wild videos captured under challenging conditions, including large head poses and diverse lighting environments.
Notably, we also include a synthetic example from URAvatar~\cite{li2024uravatar} (bottom row in \cref{fig:results_qual_wild}), which generates faces with controlled viewpoints and lighting.
Furthermore, \cref{fig:results_qual_VAT} presents examples from the gaze-following dataset VideoAttentionTarget~\cite{chong2020detecting}, a collection of diverse samples extracted from movies. 
This dataset provides annotated gaze targets, which are visualized when annotated within the image frame, as some targets may be out of frame.

These examples highlight the model's ability to predict gaze direction accurately in unseen environments, even under extreme head poses, challenging lighting conditions, and synthetic appearances.

\section{Ethical Considerations}

Our research involves the use of existing facial and gaze datasets.
In accordance with ethical guidelines, we rely on the fact that these datasets were originally collected and published following relevant ethical and data protection standards, including obtaining consent, and we do not generate or collect additional new data.
Our experimental protocols involve only image content, with no identifiable personal information or links to other personal data.

\end{document}